\documentclass[letterpaper, 10 pt, conference]{ieeeconf}  % Comment this line out if you need a4paper

\IEEEoverridecommandlockouts                              % This command is only needed if 
\usepackage{graphics} % for pdf, bitmapped graphics files

\usepackage[noadjust]{cite}

\usepackage{footnote}

\usepackage{amssymb}
\usepackage{amsmath}
\usepackage{algorithmic}
\usepackage{algorithm}
\usepackage{wrapfig}
\usepackage{booktabs}
\usepackage{lipsum}
\usepackage[font=small,labelfont=bf]{caption}
\usepackage{xcolor}
\usepackage{balance}
\usepackage{url}

\definecolor{DarkGreen}{rgb}{0.1,0.5,0.1}
\definecolor{DarkRed}{rgb}{0.5,0.1,0.1}
\definecolor{DarkBlue}{rgb}{0.1,0.1,0.5}
\definecolor{RoyalBlue}{RGB}{0,100,170}
\definecolor{SapphireBlue}{RGB}{0, 100, 200}
\definecolor{prune}{rgb}{0.44, 0.11, 0.11}
\definecolor{maroon}{rgb}{0.5450, 0, 0}
\definecolor{darkred}{rgb}{0.5450, 0, 0}
\definecolor{RoyalBlue}{RGB}{0,100,170}
\definecolor{DarkBlue}{RGB}{20,70,200}
\definecolor{peach}{rgb}{1, 0.56, 0.56}
\definecolor{NotionGreen}{RGB}{15,123,108}
\definecolor{NotionOrange}{RGB}{217,115,13}
\definecolor{NotionRed}{RGB}{224,62,62}
\definecolor{MontrealBlue}{RGB}{0, 30, 98}
\include{defs}
\def\shownotes{1}  %set 1 to show author notes
\ifnum\shownotes=1
\newcommand{\authnote}[2]{{$\ll$\textsf{\footnotesize #1: #2}$\gg$}}
\else
\newcommand{\authnote}[2]{}
\fi

\usepackage{todonotes}

\newtheorem{theorem}{Theorem}
\newtheorem{definition}{Definition}

\newcommand{\skillname}{\emph} 
\newcommand{\taskname}{\emph} 

\usepackage{selectp}
\title{\LARGE \bf
Search-Based Task Planning with Learned Skill Effect Models for Lifelong Robotic Manipulation
}

\author{
Jacky Liang$^{*1}$, Mohit Sharma$^{*1}$, Alex LaGrassa$^{1}$, Shivam Vats$^{1}$, Saumya Saxena$^{1}$, Oliver Kroemer$^{1}$% <-this % stops a space
\thanks{
$^{1}$ Robotics Institute, Carnegie Mellon University,
{\tt\small \{jackyliang, mohitsharma, lagrassa, svats, saumyas, okroemer\}@cmu.edu}
}%
\thanks{$^{*}$Equal Contribution}%
}

\begin{document}

\maketitle
\thispagestyle{empty}
\pagestyle{empty}

\begin{abstract}
Robots deployed in many real-world settings need to be able to acquire new skills and solve new tasks over time. 
Prior works on planning with skills often make assumptions on the structure of skills and tasks, such as subgoal skills, shared skill implementations, or task-specific plan skeletons, which limit adaptation to  new skills and tasks.
By contrast, we propose doing task planning by jointly searching in the space of parameterized skills using high-level skill effect models learned in simulation. 
We use an iterative training procedure to efficiently generate relevant data to train such models.
Our approach allows flexible skill parameterizations and task specifications to facilitate lifelong learning in general-purpose domains. 
Experiments demonstrate the ability of our planner to integrate new skills in a lifelong manner, finding new task strategies with lower costs in both train and test tasks.
We additionally show that our method can transfer to the real world without further fine-tuning.
\end{abstract}
\section{Introduction}
\label{sec:intro}

Lifelong-learning robots need to be able to plan with new skills and for new tasks over time~\cite{thrun1995lifelong}.
For example, a home robot may initially have skills to rinse dishes and place them individually on a rack.
Later, the robot might obtain a new skill of operating a dishwasher.
Now the robot can plan to either wash the dishes one by one or use the dishwasher depending on the costs of each skill and the number of dishes to be cleaned.
In other words, robots need to be able to obtain and use new skills over time to either adapt to new scenarios, solve new tasks, or to improve performance on existing tasks. 
Otherwise, the robot engineer would need to account for all potential tasks and strategies the robot can use before deployment.
As such, we propose a task planning system that can efficiently incorporate new skills and plan for new tasks in a lifelong robot manipulation setting.

To create such a versatile manipulation system, we use parameterized skills that can be adapted to different scenarios by selecting suitable parameter values. 
We identify three properties of skills that are important to support in this context: 1) skills can have different implementations, 2) skills can have different parameters which can take discrete, continuous, or mixed values, and 3) skill parameters may or may not correspond to subgoals.
Property one means the skills can be implemented in a variety of manners, e.g., hard-coded, learned without models, or optimized with models. 
This requires relaxing the assumptions placed on the skill structures made in previous works, such as implementing all skills with the same skill-conditioning embedding space~\cite{sharma2019dynamics, lu2020reset, xie2020skill, li2021planning}.
Property two requires the task planner to not assume any fixed structure for skill parameters.
Unlike previous works~\cite{nasiriany2019planning, mandlekar2020iris}, each skill can utilize a different number of parameters, and these parameters can be a mix of discrete and continuous values. 
% For the cleaning example, the rinsing skill may have a parameter indicating how long to rinse the dishes, while the placing skill may have a goal pose for where to place the dish.
Property three means that instead of chaining together skill subgoals, the planner needs to reason about the effects of the skills for different parameter values. 
For example, the home robot may need to predict how clean a plate is for different rinsing durations.

\begin{figure}[!t]
    \centering
    \includegraphics[width=0.9\linewidth]{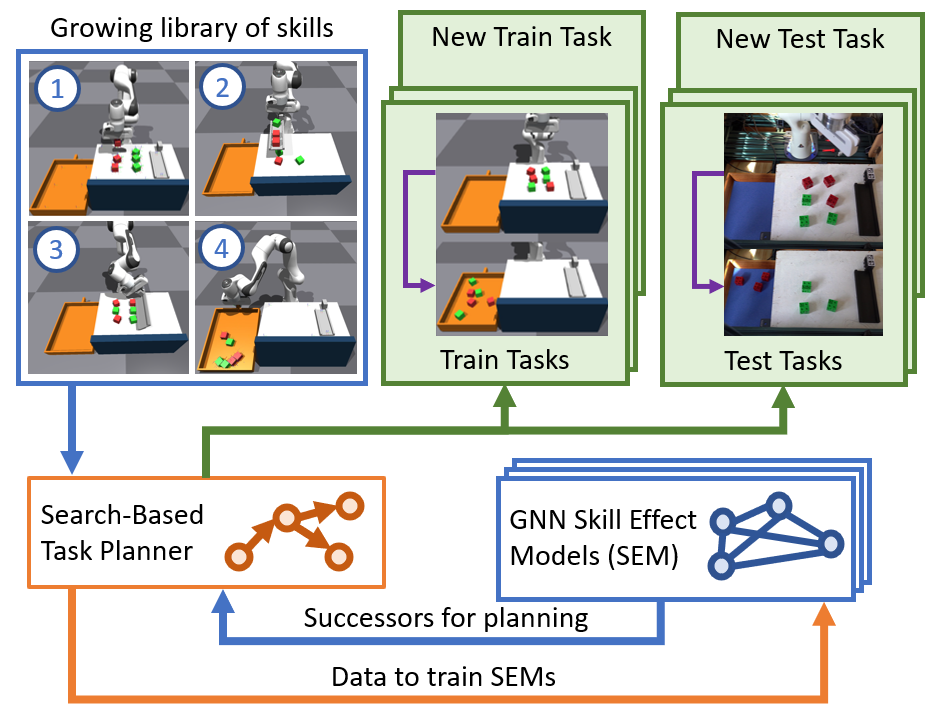}
    \caption{
        \footnotesize
        Overview of the proposed search-based task planning framework with learned skill effect models (SEMs) for lifelong robotic manipulation.
        New skills and training tasks can be added incrementally.
        We collect skill effects data by running the planner using all skills on all training tasks in simulation.
        The collected data is used to train GNN SEMs for new skills or fine-tune models of existing skills.
        Learned models predict both the terminal state and cost of skill executions.
        The planner can use SEMs to plan low-cost paths on test tasks in the real world.
        This approach supports planning 1) with a set of differently parameterized skills that can grow over time and 2) for test tasks unseen during training.
    }
    \label{fig:method}
    \vspace{-8mm}
\end{figure}

Planning for new tasks requires the planner to be flexible about the structure of task specifications.
One way to do this is by using either goal condition functions or goal distributions~\cite{conkey2020planning}, instead of shared representations like task embeddings~\cite{nasiriany2021disco} or specific goal states~\cite{srinivas2018universal, nasiriany2019planning, li2021planning, ichter2020broadly}.
Using predefined task representations limits the type of tasks a robot can do, and using learned task embeddings may require fine-tuning on new tasks.
Only having a goal condition function also makes it more difficult to represent a task as an input to a general value or policy function implemented using a function approximator.

% This approach is in contrast to previous works that specify tasks with goal states, because that implies the state representation itself can be used to encode the task.
% It may be infeasible to represent all tasks via continuous embeddings or a shared representation.
% For example, one task goal representation for the dish washing domain can be a list of final poses for each plate.
% This representation supports a class of tasks, each with a different combination of final poses.
% However, this shared representation is inflexible about small but practical changes in the task.
% The user can specify a task that is achieved when all plates are cleaned and placed on a drying rack, where specific plate poses are not important, so long as the plates plates are not placed too closely to one another.
% In this case, the final goal pose of each plate is underspecified by the task, so it cannot be directly described with a set of goal poses.
% While describing the task with one set of goal poses is possible, doing so in this case would prevent the planner from finding potentially lower-cost solutions, and doing so in general might not be feasible.

To satisfy the skill and task requirements for the lifelong manipulation planning problem, we propose a task planning system that performs \textbf{search-based planning with learned effects of parameterized skills}.
Search-based methods directly plan in the space of skill-parameter tuples.
A key advantage of search-based planning methods is they can use skills regardless of parameter choices or implementation details, and only need a general goal condition check to evaluate task completion.
% Additionally, these methods provide completeness and suboptimality guarantees, which we show for our method in section~\ref{subsec:planning}.

To efficiently use search-based planning methods for task planning, we propose to learn skill effect models (SEMs).
SEMs are learned instead of hardcoded or simulated, since manually engineering models is not scalable for complex skills and simulations are too expensive to perform online during planning.
Every skill has its own SEM that predicts the terminal state and costs of a skill execution given a start state and skill parameters.
% Cost prediction allows the planner to find low-cost solutions.
% SEMs are trained with transition generated by the planner while planning on a set of training tasks.
We interleave training SEMs with generating training data by running the planner with the learned SEMs on a set of training tasks.
Our data collection method efficiently collects skill execution data relevant for planning, and supports the addition of new skills and tasks over time.
The planner uses the SEMs to plan for existing tasks with different initial states, as well as new test tasks.

Our contributions are 1) a search-based task planning framework with learned skill-effect models that 2) relaxes assumptions of skill and task representations in prior works;
skill effect models are learned with 3) an iterative data collection scheme that efficiently collects relevant training data, and together they enable 4) planning with new skills and tasks in a lifelong manner.
Please see supplementary materials, with additional results and experiment videos, at \url{https://sites.google.com/view/sem-for-lifelong-manipulation}.

\section{Related Works}
\label{sec:rw}

\textbf{Subgoal skills.}
Many prior works approached planning with skills with the subgoal skill assumption. 
The successful execution of a subgoal skill always results in the same state or a state that satisfies the same preconditions of all skills, regardless of where the skill began in its initiation set~\cite{konidaris2018skills}.
As such, the skill effects are always known, and such approaches instead focus on learning preconditions~\cite{nasiriany2019planning} of goal-conditioned policies, efficiently finding parameters that satisfy preconditions~\cite{mandlekar2020iris, simeonov2020long}, or learning feasible skill sequences~\cite{ichter2020broadly}.
While subgoal planning is powerful, it limits the types of skills the robot can use.

% In~\cite{nasiriany2019planning}, the authors learn a continuous scoring function to predict the preconditions of low-level goal-conditioned policies, and their planner uses the Cross Entropy Method (CEM) to optimize over a sequence of subgoals that maximize this precondition score.
% Instead of learning preconditions directly, the authors of~\cite{simeonov2020long} learn a subgoal parameter sampler from successful skill executions.
% Planning with the learned parameter sampler is more efficient than hardcoded samplers, but the method requires knowledge of plan skeletons --- a sequence of skills without parameters --- and the planner only optimizes for skill parameters, not sequences of (skill, parameter) tuples.

% Planning with subgoal skills can also forgo the notion of preconditions by learning goal-conditioned skills and always assuming that the skill reaches the goal.
% In~\cite{ichter2020broadly}, the authors use a sampling-based planning strategy inspired by Rapidly Exploring Random Trees (RRT) to plan over sequences of subgoals for learned goal-conditioned policies without explicit precondition checks.
% The method in~\cite{mandlekar2020iris} learns goal-conditioned policies, a task-specific goal sampler, and a task-specific value function.
% During task execution, the high-level policy samples a set of subgoals and follows the one with the best predicted value --- no planning required.
% In both works learning is done through demonstrations.

\textbf{Non-subgoal skills.}
For works that plan with non-subgoal skills, many represent the skill policy as a neural network that takes as input both the state and an embedding that defines the skill.
This can be viewed as planning with one parameterized skill or a class of non-parameterized skills, each defined by a different embedding.
Such skills can be discovered by experience in the real world~\cite{lu2020reset} and in learned models~\cite{sharma2019dynamics, xie2020skill}, or learned from demonstrations~\cite{li2021planning}.
Planning with these skills is typically done via Model Predictive Control (MPC), where a short sequence of continuous skill embeddings is optimized, and replanning occurs after every skill execution.
While these approaches do not assume subgoal skills, they require skills to share the same implementation and space of conditioning embeddings, and MPC-style planning cannot easily support planning with multiple skills with different parameter representations~\cite{nasiriany2019planning, xie2020skill, lu2020reset, li2021planning}.

% In~\cite{lu2020reset} the authors propose a method to automatically discover such skills with intrinsic rewards that encourage skills with different condition embeddings to be predictable and distinguishable.
% Skill policies, embeddings, and a skill-level dynamics model are learned by exploration in the real world.
% To achieve a task, Model Predictive Control (MPC) is used to optimize a sequence of skill condition embeddings given a task-specific reward function.
% The algorithms in~\cite{sharma2019dynamics, xie2020skill} are similar, but they allow skill discovery using the learned dynamics model, instead of real-world rollouts.
% In~\cite{sharma2019dynamics} the authors also improve learning efficiency by automatically generating a curriculum to train the discovered skills.
% The method in~\cite{li2021planning} also plans with latent-conditioned skills, but here the skills are learned via imitation learning instead of discovered.

\textbf{Planning with parameterized skills.}
To jointly plan sequences of different skills and parameters, works have proposed a two-stage approach, where the planner first chooses the skills, then optimizes skill parameters~\cite{pan2020decision, xu2020deep, simeonov2020long}.
% One example is~\cite{pan2020decision}, which has model-based skills that operate in the state space and obtains skill effects via simulation.
% Another is~\cite{xu2020deep}, which learns latent dynamics and latent preconditions of hardcoded skills along with a model that proposes plan skeletons.
Unlike directly searching with skills and parameters, it is difficult for two-stage approaches to give guarantees on solution quality.
Some also require hardcoded or learned plan skeletons~\cite{xu2020deep, simeonov2020long}, which limits the planner's applicability to new tasks.

Instead of planning, an alternative approach is to learn to solve Markov Decision Processes (MDPs) with parameterized skills~\cite{masson2016reinforcement, hausknecht2015deep, xiong2018parametrized}.
However, learning value or policy functions typically requires a fixed representation for function approximators, so these methods cannot easily adapt to new skills and skills with parameters with different dimensions or modalities (e.g. mixed continuous and discrete).
Doing so for search-based planning can be done by directly appending new skills when expanding a node for successors.

\textbf{Obtaining skill effects.}
Many prior works used simulated skill outcomes during planning~\cite{kim2017parts, song2019multi, kim2019pomhdp, pan2020decision}.
This can be prohibitively expensive to perform online, depending on the complexity of simulation and the duration of each skill.
To avoid simulation rollouts, works have used hardcoded analytical~\cite{bagaria2020skill, butzke2014state} or symbolic~\cite{kaelbling2011hierarchical, kaelbling2013integrated, eppe2019semantics} skill effect models.
Manually engineering such models may not always be feasible, and they do not easily scale to changes in skills, dynamics, and tasks.
Although symbolic models can be automatically learned~\cite{ugur2015bottom, konidaris2018skills, ames2018learning, suarez2020leveraging, wang2021learning}, these approaches also make the subgoal skill assumption.
By contrast, our method, which learns skill effect models in continuous states without relying on symbols, can plan with both subgoal skills as well as skills that do not share this property.

The works most closely related to ours are~\cite{xu2020deep} and~\cite{wang2021learning}.
In~\cite{xu2020deep}, the authors jointly train latent dynamics, latent preconditions, and parameter samplers for hardcoded skills and a model that proposes plan skeletons.
Planning is done MPC-style by optimizing skill parameters with the fixed plan skeleton.
Although this approach does not assume subgoal skills and supports skills with different parameters, learning task-specific plan skeletons and skill parameter samplers makes it difficult to use for new tasks without finetuning.
% There is also the added complexity of latent space training.
% Since all learned models must share the same latent space, the addition of new skills would fine-tuning all models jointly.
The method in~\cite{wang2021learning} learns to efficiently sample skill parameters that satisfy preconditions.
Task planning is done using PDDLStream~\cite{garrett2020pddlstream}, which supports adding new skills and tasks.
Though this approach does not use subgoal parameters, the desired skill outcomes are narrow and predefined, and the learned parameter sampler aims to achieve these predefined effects.
As such, the method shares the limitations of works with subgoal skills, where the skill-level transition model is not learned but predefined as the subgoals.
\section{Task Planning with Learned Skill Effect Models}
\label{sec:method}

% \subsection{Overview}

% \begin{figure*}[!t]
%     \centering
%     \includegraphics[width=0.8\linewidth]{ims/method_overview.png}
%     \caption{
%         \footnotesize
%         Overview of the proposed search-based task planning framework with learned skill effect models for lifelong robotic manipulation.
%         New skills and new training tasks can be added incrementally.
%         We collect skill effects data by running the planner using all skills on all training tasks in simulation.
%         The collected data is used to train GNN skill effect models for new skills or fine-tune models for existing skills.
%         Learned models predict both the terminal state of skill execution as well as execution cost, which enables search-based task planning that finds low-cost paths.
%         The planner can use the learned models to plan on test tasks in the real world.
%         This approach supports planning 1) with a set of differently parameterized skills that can grow over time and 2) for test tasks unseen during training.
%     }
%     \vspace{-20pt}
%     \label{fig:method}
% \end{figure*}

The proposed method consists of two main components - learning skill effect models (SEMs) for parameterized skills and using SEMs in search-based task planning.
These two components are interleaved together - we run the planner on a set of training tasks using SEMs to generate data, which is used iteratively to further train the SEMs.
New skills and training tasks can be added to the pipeline because the planner and the SEMs do not assume particular implementations of skills and tasks.
The planner can also directly apply the learned SEMs to solve test tasks.
See overview in Figure~\ref{fig:method}.

\subsection{Skill Planning Problem Formulation}

\textbf{Parameterized skills.}
Central to our approach is the options formulation of skills~\cite{sutton1999between, konidaris2018skills}.
Denote a parameterized skill as $o$ with parameters $\theta \in \Theta$.
Parameters are skill-specific and may contain subgoal information such as the target object pose for a pick-and-place skill.
We assume a fully observable state $x \in \mathcal{X}$ that contains all information necessary for task planning, cost evaluations, and skill executions.
% Some information contained in $\mathcal{X}$ might not be useful for some skills, and we assume it is sufficient for all possible skills.
We define the low-level action $u \in \mathcal{U}$ as the command sent to the robot by a low-level controller shared by all skills (e.g. torque).

In our formulation, a parameterized skill $o$ contains the following $5$ elements:
an initiation set (precondition) $\mathcal{I}_o(x, \theta) \to \{0,1\}$, a parameter generator that samples valid parameters from a distribution $p_o(\theta | \mathcal{I}(x, \theta_i) = 1)$, a policy $\pi_o(x) \to u$, a termination condition $\beta_o(x, \theta, t) \to \{0, 1\}$, and the skill effects $f_o(x_t, \theta) \to x_{t+T}$, where $T$ is the time it took for the skill to terminate.
To execute skill $o$ at state $x$ with parameters $\theta$, we first check if $(x, \theta)$ satisfies the preconditions $\mathcal{I}_o$.
If it does, then we run the skill's policy $\pi_o$ until the termination condition is satisfied.
We assume that the preconditions, parameter generator, policy, and termination conditions are given, and the skill effects are unknown but can be obtained by simulating the policy.
To enable reasonable planning speeds, the SEMs learn to predict these skill-level transitions.

To justify the assumption of given skill preconditions, we note that our preconditions are broader than ones in prior works and consequently can be easily manually defined.
Preconditions in many prior works, especially ones that use subgoal skills, are only satisfied when a specific outcome is reached, so they may require learning sophisticated functions to classify which (state, parameter) tuple lead to the intended outcome~\cite{xu2020deep, wang2021learning}.
By contrast, because we allow non-subgoal skills, our preconditions are satisfied if skill execution leads to any non-trivial and potentially desirable outcome.
For example, for a table sweeping skill, the preconditions are satisfied as long as the robot sweeps something, instead of requiring sweeping specific objects into specific target regions.
Due to the broad and simple nature of our more flexible preconditions, we argue it is reasonable to assume they are given.

\textbf{Task planning of skills and parameters.}
Before specifying tasks, we first define a background, task-agnostic cost $c(x_t, u_t) \ge 0$ that should be minimized for all tasks.
% This cost function may contain terms that penalize action norms (e.g. excessive torques), risky states (e.g. being too close to obstacles), or time (e.g. skills that take many time steps incur higher costs).
This cost is accumulated at each step of skill $o$ execution, so the total skill cost is $c_o = \sum_{t=0}^T c(x_t, u_t)$.
% The SEMs also predict this total skill execution cost to enable cost-aware planning.
A task is specified by a goal condition $\mathcal{G}(x)\to \{0, 1\}$ that classifies whether or not a state achieves the task.
We denote a sequence of skills, parameters, and their incurred states as a path $P = (x_0, o_0, \theta_0, x_1, \hdots x_n, o_n, \theta_n, x_{n+1},  \hdots, x_N)$, where $N$ is the number of skill executions, and the subscripts indicate the $n$th skill in the sequence (not time).
We assume the environment dynamics and skill policies are deterministic.
The task planning problem is to find a path $P$ such that the goal condition is satisfied at the end of the last skill, but not sooner, and the sequence of skill executions is feasible and valid.
See equation~\ref{eq:task_plan}.
\begin{align} \label{eq:task_plan}
           \min_P \quad & \sum_{n=0}^{N-1} c_{o_n} \\
    \textrm{s.t.} \quad & \mathcal{G}(x_N) = 1  \notag \\
    \forall n\in [0, N-1], & \mathcal{G}(x_n) = 0 \notag \\
    \mathcal{I}_{o_n}(x_n, \theta_n) = 1, & f_{o_n}(x_n, \theta_n) = x_{n+1} \notag
\end{align}
Note that $\theta$, $\mathcal{I}_o$, and $f_o$ are all skill-specific, so with $M$ types of skills, there are $M$ different parameter spaces, preconditions, and skill effects.

\subsection{Learning Skill Effect Models (SEMs)}

\textbf{Defining SEMs for manipulation skills.}
We learn a separate SEM for each skill, which takes as input the current state $x_t$ and a skill parameter $\theta$.
The SEM predicts the terminal state $x_{t+T}$ reached by the skill when it is executed from $x_t$ using $\theta$ and the total skill execution cost $c_o$.
We assume SEMs are queried only with state and parameter tuples that satisfy the precondition.
Because we focus on the robot manipulation domain, we assume the state space $\mathcal{X}$ can be decomposed into a list of object-centric features that describe discrete objects or robots in the scene.
% As such, our SEMs additionally predict an object-level mask that signifies whether or not an object is affected by the skill execution.
% During planning time, only features of objects that are predicted to be affected by the skill are updated with the network predictions.
% This helps make sequential SEM queries more robust, as neural network predictions are never exact and can lead to static objects seemingly drifting over multiple predictions.

We represent SEMs using Graph Neural Networks (GNNs), because their inductive bias can efficiently model interactions among entities through message passing, encode order-invariance, and support different numbers of nodes and edges during training and testing~\cite{battaglia2016interaction, janner2018reasoning, kartmann2018extraction, tekden2021object}.
Each node in the SEM GNN corresponds to an object in the scene and contains features relevant to that object from the state $x$.
We denote these object features as $s_k\in\mathbb{R}^S$, where $k$ denotes the $k$th object in the scene.
Because a skill may directly affect multiple objects, each node also contains the skill parameters $\theta$ as additional node features. 
The full node feature is the concatenation of $[s_k, \theta]$.
There are no edge features.
The network makes one node-level prediction, the change in object features $\Delta s_k$, and one graph-level prediction, the total skill execution cost $c_o$.
% the probability an object is affected by the skill execution $p_k = p(d_k | s_k, \theta)$, where $d_k$ is the binary affected label, and 
As SEMs make long-term predictions about the entire skill execution, the graph is fully connected to allow all objects to interact with each other, not just objects that are initially nearby.
The loss function to train SEMs for a single step of skill execution prediction is 
$\mathcal{L} = \lambda_c\|c_o - \hat{c_o}\|_2^2 + \frac{\lambda_s}{K}\sum_{k=1}^K \|\Delta s_k - \hat{\Delta s_k} \|_2^2$.
%- \lambda_p\left(d_k\log \hat{p_k} + (1 - d_k)\log(1 - \hat{p_k})\right)
The hat notation denotes predicted quantities, and the $\lambda$s are positive scalars that tune the relative weights between the loss terms.
The GNN is implemented with PyTorch Geometric~\cite{Fey/Lenssen/2019}.

SEMs enable efficient planning of diverse parameterized skills, as well as two additional benefits.
% First, using learned SEMs speeds up planning by avoiding expensive simulation rollouts.
First, because the model is on the skill-level, not action-level, it only needs one evaluation to predict the effects of a skill execution, which reduces planning time as well as covariate shift by reducing the number of sequential predictions~\cite{seker2019deep, naderian2020c, janner2020gamma, lambert2020learning}.
Second, a long-horizon skill-level model can leverage a skill's ability to act as a funnel in state space during execution, which simplifies the learning problem.

\textbf{Collecting diverse and relevant data for training SEMs.}
To learn accurate and generalizable SEMs, they must be trained on a set of skill execution data that is both diverse and relevant to task planning.
While we assume knowledge of the initial state distribution of all tasks, we do not know the distribution of all states visited during planning and execution.
As we cannot manually specify this incurred state distribution, we obtain it and train the SEMs in an iterative fashion that interleaves SEM training with data generation by planning and execution, as seen in Figure~\ref{fig:method}.
First, given an initial set of skills, we generate single skill execution transitions from the known initial state distribution.
This data is used to train the initial SEMs.
Then, given a set of training tasks, we use the planner to plan for these tasks using the learned SEMs across a set of initial states.
The planner terminates when it finds a path to the goal or reaches a fixed planning budget (reaching maximum number of nodes expanded, maximum search depth, or maximum planning time).
% Then, we trace a set of paths from the root planning node (initial state) to nodes that either achieve the goal condition or are the closest to the goal.
Then we sample paths in the graph and simulate them to collect skill execution data, which is added to a dataset of all skill data collected so far.
Path sampling is biased toward longer paths and ones that have the newly added skills.
The transitions added are filtered for duplicates, since multiple paths in a planning graph may share the same initial segments which would bias the dataset towards transitions closer to the initial states.
After a fixed amount of path data is collected, we continue training the SEMs on the updated dataset before restarting the data collection process.
In the beginning, it is expected that the planner performance will be highly suboptimal due to the inaccurate initial SEMs.
While we use simulation data due to benefits in speed, this is not a requirement and SEMs can be trained with real-world data.
% Collecting data in simulation is faster than doing so in the real-world, which must run at real-time, cannot be easily parallelized, and requires humans or complex mechanisms for setting states.
% See summary in Algorithm~\ref{alg:iterative_training}.

\textbf{Planning with new skills.}
The above procedure supports incrementally expanding the list of skills used by the planner.
Given a new skill, we first train an initial SEM by sampling from the initial state distribution, then during planning data generation the search-based planner can use the new SEM to get successors.
SEMs for new and existing skills will be improved and continuously trained on this new planning data.
Fine-tuning previous SEMs is needed, because the new skill might have incurred states that were previously absent from the dataset.
Although this fine-tuning may not be necessary in specific cases, we leave detecting such scenarios and reducing overall training budget to future work.
Learning one SEM for each skill allows for different parameter spaces (e.g. dimensions, discrete, continuous, mixed) that cannot be easily represented with a shared, common model.

\textbf{Planning with new tasks.}
Because the planner does not rely on predefined plan skeletons, it can directly use SEMs to plan for new tasks.
Two main factors about data collection affect the generalization capability of the SEMs when applied to unseen test tasks.
The first is whether the states incurred while planning for training tasks are sufficiently diverse and relevant to cover the states incurred by planning for test tasks.
The second is the planner itself --- how greedy is its search and how much it explores the state space.
Many planners have hyperparameters that can directly balance this exploration-exploitation trade-off.
% , balancing the data generation speed, performance on training tasks, and range of the generated data.

\subsection{Search-based Task Planning}
\label{subsec:planning}

We pose task planning as a graph search problem over a directed graph, where each node is a state $x$, and each directed edge from $x$ to $x'$ is a tuple $(o, \theta)$ such that $f_o(x, \theta) = x'$.
Edges also contain the costs of skill executions $c_o$.
During search, this graph is constructed implicitly.
Given a node to expand, we iterate over all skills, generate up to $B_o$ parameters per skill that satisfy the preconditions, then evaluate the skill-level dynamics on all state-parameter tuples to generate successor states.
$B_o$ decides the maximum branching factor on the graph.
This number varies per skill, because some skills have a broader range of potential parameters than others, requiring more samples.
% The number of parameters actually sampled could be $0$ if no parameters satisfy the precondition of the skill at the given state.
% It could also be a number between $0$ and $B_o$ if a maximum sampling budget is reached for rejection sampling with the preconditions.

To search on this graph, we apply Weighted A* (WA*), which guarantees completeness on the given graph.
If the heuristic is admissible, WA* also guarantees the solution found is no worse than $\epsilon c^*$, where $c^*$ is the cost of the optimal path and $\epsilon$ determines how greedily the search follows the heuristic.
We assume an admissible heuristic is given.
This is in line with previous works that have shaped rewards or costs that guide the planner~\cite{nasiriany2019planning, xu2020deep, lu2020reset, li2021planning}.

% While WA* provides guarantees on a graph, we additionally need to show the constructed graph represents the underlying problem sufficiently well.
% % The choice of $B_o$ is critical in this respect.
% Under smoothness assumptions in the dynamics and cost functions, with large but bounded $B_o$s the graph will contain a solution that is close to the optimal with high probability.
% % This relationship is a function of (1) the smoothness of the skill dynamics and the cost function (2) length of the optimal solution and (3) the goal set.
% A detailed analysis is provided in Appendix~\ref{app:planner}.

The proposed method enables planning with new skills and to solve new tasks in continuous states.
Planning for new tasks is done by replacing the heuristic and goal conditions, which does not affect the graph construction procedure or the SEMs.
Searching in continuous states is more flexible than searching in symbolic states, and it is not necessarily slower.
Flexibility comes from the ability to integrate new skills and tasks without needing to create new symbols. 
Planning speed depends on the size of the action space (branching factor) and the state space.
Using symbolic instead of continuous states does not reduce the branching factor, and partitioning continuous states into symbolic states without subgoal skills yield little benefits~\cite{konidaris2018skills}.
\section{Experiments}
\label{sec:exps}

\begin{figure}[!t]
    \centering
    \vspace{5pt}
    \includegraphics[width=1\linewidth]{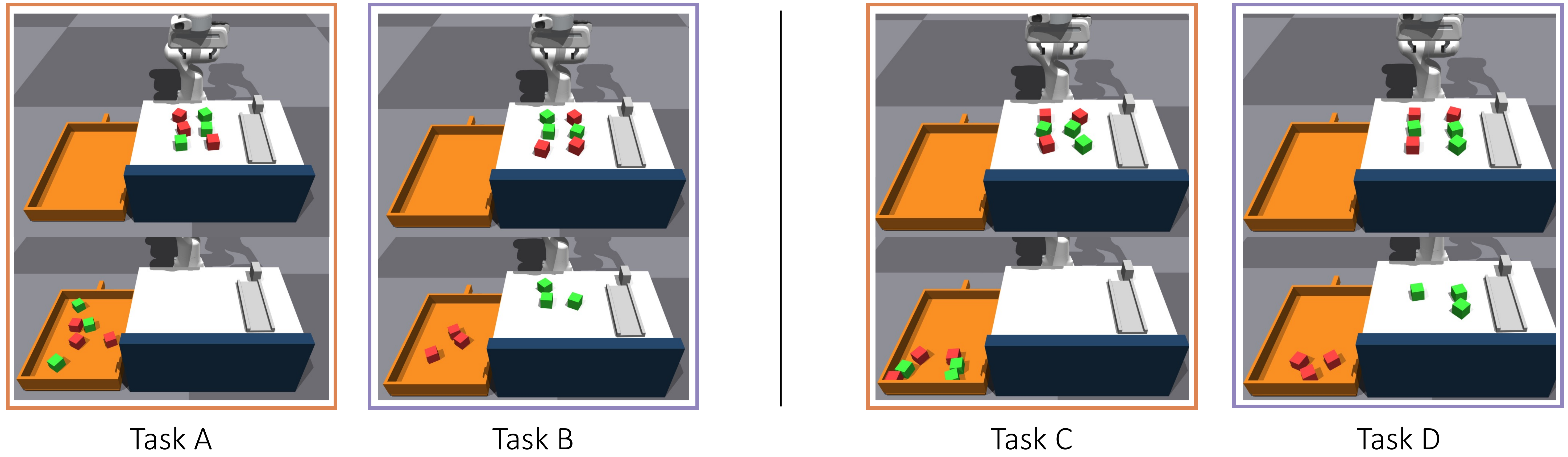}
    \caption{
        \footnotesize
        Different tasks used in our experiments. 
        The top row shows examples of initial states, the bottom shows examples of goal states.
        \emph{Left:} blocks to bin tasks (tasks \emph{(A,B)}). \emph{Right:} blocks to far bin tasks (tasks \emph{(C,D))}.
    }
    \label{fig:task_images}
    \vspace{-12pt}
\end{figure}

% \begin{figure}[!t]
%     \centering
%     \includegraphics[width=0.9\linewidth]{ims/skills.png}
%     \caption{
%         \footnotesize
%         The four skills in the blocks and bin manipulation domain.
%         We evaluate our algorithm by incrementally integrating new skills over time and measure how their additions affect task planning performance.
%     }
%     \label{fig:skills}
% \end{figure}

% \Mnote{
% 1) How efficiently can we add different skills (subgoal+no-subgoal) in our framework?
% 2) Can we use the learned SEMs on test tasks?
% 3) Planning speed, optimal plans?
% }
Our main experiment analyzes the effect of incrementally adding new skills to the proposed method on planning performance of both train and test tasks.
We apply our method to a block manipulation domain (Figure~\ref{fig:task_images}) because it can be reliably simulated, contains a diverse set of skills, and the skills have broader applications in desktop manipulation and tool use.
In addition, we show our approach compares favorably against planning with simulation and the benefits of using planning data to train SEMs.
% To demonstrate additional benefits of our approach we compare it against planning with simulations, 
%latent-space dynamics models, 
%and training SEMs with random data instead of planning data.
Lastly, we show the generalizability of our method by deploying it in a real-world setup.
More experiment details are in Appendix-\ref{app:exp-details}.

\subsection{Task Domain}

The task domain has a Franka Emika Panda 7 DoF arm, a set of colored blocks, a table, a tray, and a bin.
On the table, blocks of the same size and different colors are initialized in random order on a grid with noisy pose perturbations. 
The tray on the table can be used as a tool to carry and sweep the blocks.
Beside the table is a bin, which is divided into two regions, the half which is closer to the robot, and the half that is farther away.
The state space contains the 3D position of each block, color, and index.
We implement the task domain in Nvidia Isaac Gym~\cite{ig}, a GPU-accelerated robotics simulator~\cite{liang2018gpu} that enables fast parallel data collection.

\textbf{Skills.} 
We experiment with four skills:
\skillname{Pick and Place} (Figure~\ref{fig:method} skill 1) moves a chosen block to a target location.
It has a mixed discrete and continuous parameter space --- which object to pick and its placement location.
\skillname{Tray Slide} (Figure~\ref{fig:method} skill 2) grasps the tray, moves it to the bin, and tilts it down, emptying any blocks on it into the bin. 
Its parameter is a continuous value defining where along the length of the bin to rotate the tray.
\skillname{Tray Sweep} (Figure~\ref{fig:method} skill 3) uses the tray to perform a sweeping motion along the table.
Its parameter specifies where to start the sweeping motion, and the sweep motion ends at the table's edge.
\skillname{Bin Tilt} (Figure~\ref{fig:method} skill 4) grasps the handle at the side of the bin and tilts the bin by lifting the handle, which moves blocks in the bin from the close half to the far half.
Skills are implemented by following open-loop trajectories defined by the skill parameters.
We did not learn more complex skills as our work focuses on task planning and not skill learning.

\textbf{Tasks.}
We evaluate on four different tasks (Figure~\ref{fig:task_images}) that are variations of moving specific sets of blocks to different regions in the bin.
Two tasks are used to collect SEM training data: \taskname{Move All Blocks to Bin (A)} and 
\taskname{Move All Blocks to Far Bin (C)},
while the remaining two are used to evaluate learned SEMs: \taskname{Move Red Blocks to Bin (B)} and \taskname{Move Red Blocks to Far Bin (D)}.
Each task uses the same background cost function, which is the distance the robot's end-effector travels, plus a small penalty for placing the gripper inside the bin.
The admissible heuristic used is the mean distance of each block to the closest point in their target regions. 

While \skillname{Pick and Place} can make substantial progress on all tasks, it alone is not sufficient because kinematic constraints inhibit the robot from directly placing blocks on the far side of the bin, so \skillname{Bin Tilt} or \skillname{Tray Slide} is needed.
Additionally, using other skills can achieve lower costs; \skillname{Tray Sweep} can quickly move multiple blocks into the bin, but this may move blocks that need to stay on the table.
The sequence of skills may change depending on the initial placement of the blocks, and the path needs to be low-cost.

% \textbf{Metrics.} 
% First, we use execution costs to verify if planning with SEMs can find optimal plans.
% Second, we evaluate how the proposed method enables lifelong integration of new skills by recording
% the probability of finding optimal plans over time with increasing amount of training data.
%since we train SEMs iteratively on the planner data we also evaluate the probability of finding optimal plans using SEMs with increasing amount of planning data.
% Finally, we also evaluate the plan times using SEMs.

\subsection{Lifelong Task Planning Results}

\begin{figure}[t]
    \centering
    \vspace{6pt}
    \includegraphics[width=0.8\linewidth]{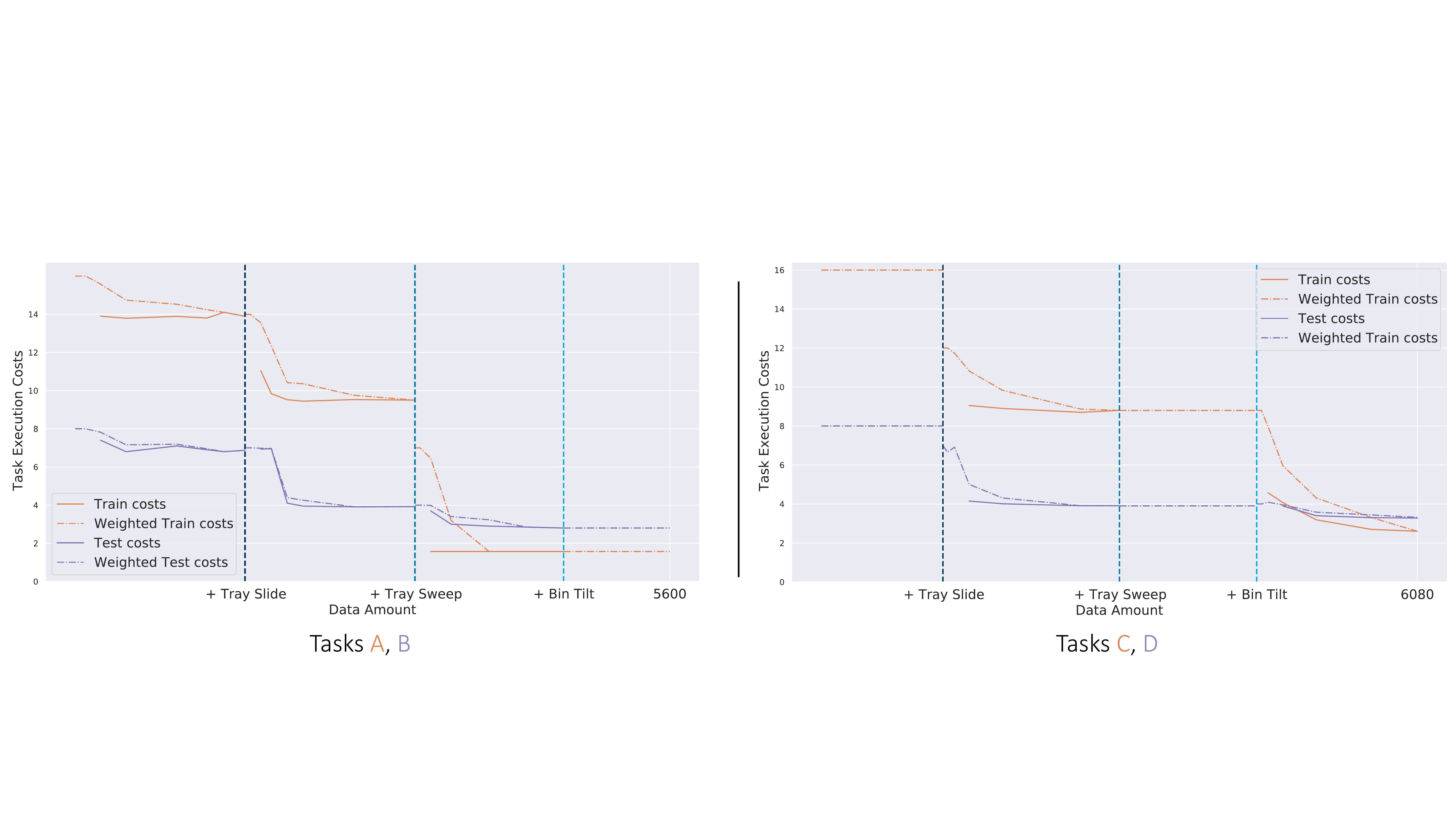}
    \includegraphics[width=0.8\linewidth]{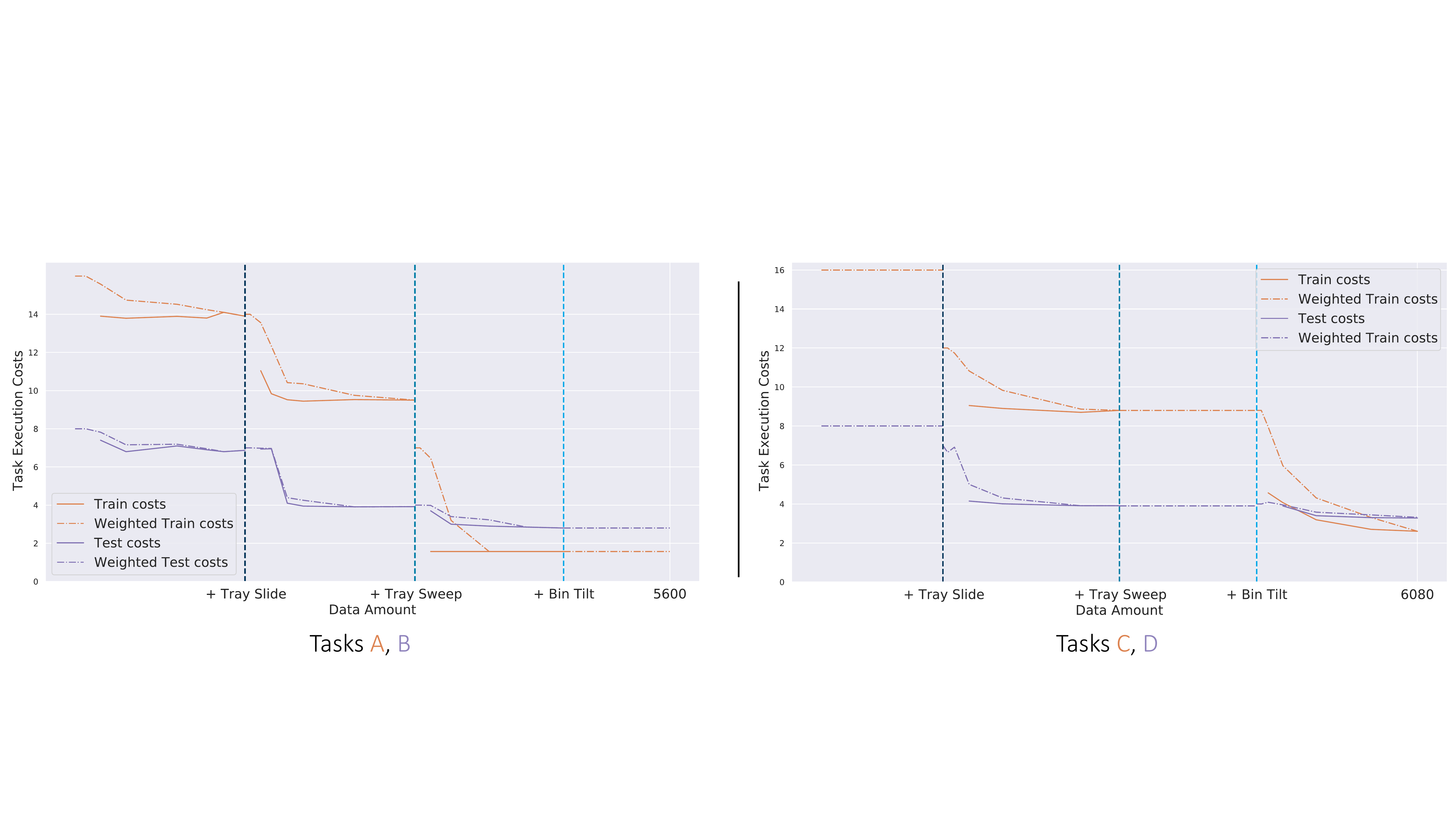}
    \caption{
        \footnotesize
        Task execution costs plotted over time as new skills are learned and integrated in a lifelong manner.
        Blue vertical lines signify the addition of a new skill.
        Weighted costs are calculated by weighting the task cost with the success rate.
    }
    \label{fig:costs_all_tasks}
    \vspace{-10pt}
\end{figure}

\begin{figure}[t]
    \centering
    \includegraphics[width=0.8\linewidth]{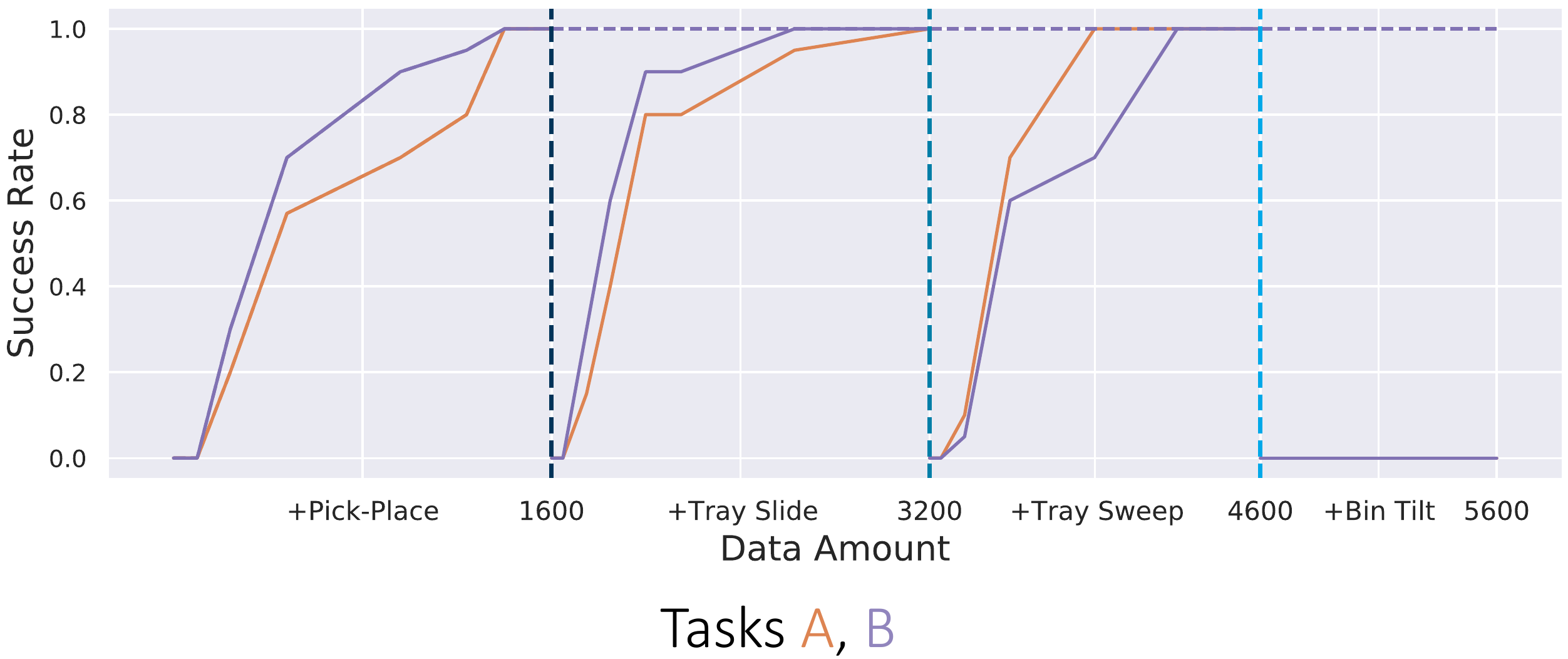}
    \includegraphics[width=0.8\linewidth]{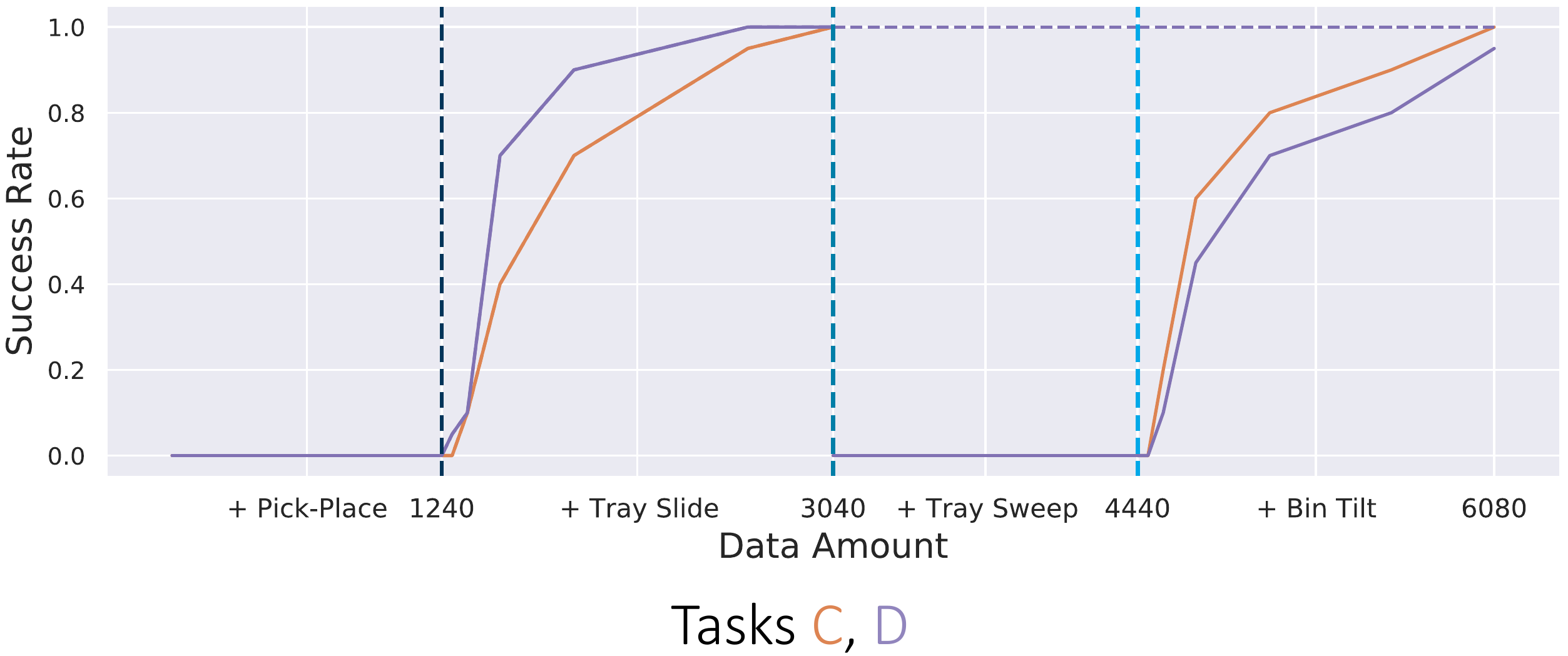}
    \caption{
        \footnotesize
        Task execution success rate for each new added skill. 
        Each skill is being added over time.
        Orange are train tasks; purple are test tasks.
        Solid lines are planning with new skills; dashed are with any skills.
    }
    \label{fig:succ_prob_all_tasks}
    \vspace{-7mm}
\end{figure}

To evaluate our approach for lifelong integration of new skills, we add the four skills over time using the iterative training procedure. 
We evaluate two scenarios, first in which the train-test task pair are respectively tasks A and B, and second with C and D.
In each case, the robot starts with only \skillname{Pick and Place}, while \skillname{Tray Slide}, \skillname{Tray Sweep}, and \skillname{Bin Tilt} are added successively in that order at pre-determined intervals.
We measure planning performance using execution costs, execution success, and planning time.
For each goal, the robot plans only once from the initial state, which terminates when it succeeds or times out.

Figure~\ref{fig:costs_all_tasks} plots the execution costs over time for both scenarios.
The proposed method is able to incorporate new skills over time, lowering execution costs when applicable by planning with new skills.
For example, adding \skillname{Tray Slide} allows the planner to find plans with significantly reduced costs across all tasks, since multiple blocks can now be moved together.
In other cases, adding a new skill does not affect task performance.
% , and the planner is able to find previously discovered plans and maintain the same execution costs.
One example is adding \skillname{Bin Tilt} to the blocks to anywhere in bin tasks \emph{(A,B)}, because the main use of the skill is to move blocks to the far side of the bin.
Another is on adding \skillname{Tray Sweep} --- it significantly reduced costs for moving all blocks to the bin \emph{(A,C)}, but less so for moving only red blocks to the bin \emph{(B,D)}.
This is because sweeping is only useful for the latter task when multiple red blocks line up in a column near the bin, which rarely occurs in the randomly initialized states. 

Figure~\ref{fig:succ_prob_all_tasks} plots the success rate of finding successful plans (dashed) and optimal plans (solid) with new skills.
Immediately after adding a new skill, there is insufficient data to learn a robust SEM, so the planner is unlikely to find optimal plans using the new skill.
Or, if it does find a plan, the plan often leads to execution failures.
As more data is collected, SEM accuracy improves and the probability of finding optimal plans increases.
Figure~\ref{fig:succ_prob_all_tasks} also shows how some tasks can only be completed after a new skill is incorporated.
For instance, with just \skillname{Pick and Place}, the robot can accomplish blocks to bin tasks \emph{(A,B)}, but fails to plan for the blocks in far bin tasks \emph{(C,D)}.
Adding new skills for \emph{(A,B)} did not change the success rate of the task, which remained at 100\%, although the composition of the plans found does change.
For \emph{(C,D)}, adding \skillname{Tray Slide} enabled 100\% success rate, while adding \skillname{Tray Sweep} did not affect plan compositions, but adding \skillname{Bin Tilt} did. 
These results show that our proposed method can learn skill effects and plan with SEMs in a lifelong manner, and that SEMs can plan for new tasks without additional task-specific learning. Qualitative results can be found in Appendix~\ref{app:additional-results}.

\begin{table}[!t]
    \centering
    \vspace{5pt}
    \begin{tabular}{@{}lll@{}}
        \toprule
        Task & Sim            & SEMs (Ours)        \\ \midrule
        A    & 776.19 (46.9) & 1.3 (0.7) \\
        C    & 1736.8 (187.) & 0.98 (0.3) \\ \bottomrule
    \end{tabular}
    \caption{
        \footnotesize
        Comparing plan times in seconds using simulator vs. SEMs.
        Parenthesis indicate standard deviations.
    }
    \label{tab:learn_model_vs_sim_quantitative}
\end{table}

\begin{table}[!t]
    \centering
    \begin{tabular}{@{}lllll@{}}
        \toprule
        Task   & \footnotesize{Pick-Place} & \footnotesize{+Tray-Slide} & \footnotesize{+Tray Sweep} & \footnotesize{+Tilt Bin}  \\ \midrule
        A & 11.3 (3.4)   & 20.2 (7.9)   & 0.6 (0.5)    & 1.3 (0.7)   \\
        B & 7.4 (2.3)    & 14.9 (8.2)   & 18.0 (14.3)  & 22.1 (12.4) \\ \bottomrule
    \end{tabular}
    \caption{
        \footnotesize
        Plan times (seconds) using SEMs for objects to bin tasks (A, B) with an increasing number of skills.
    }
    \label{tab:sem_plan_time}
    \vspace{-5pt}
\end{table}

\begin{figure}[!t]
    \centering
    \includegraphics[width=0.3\textwidth]{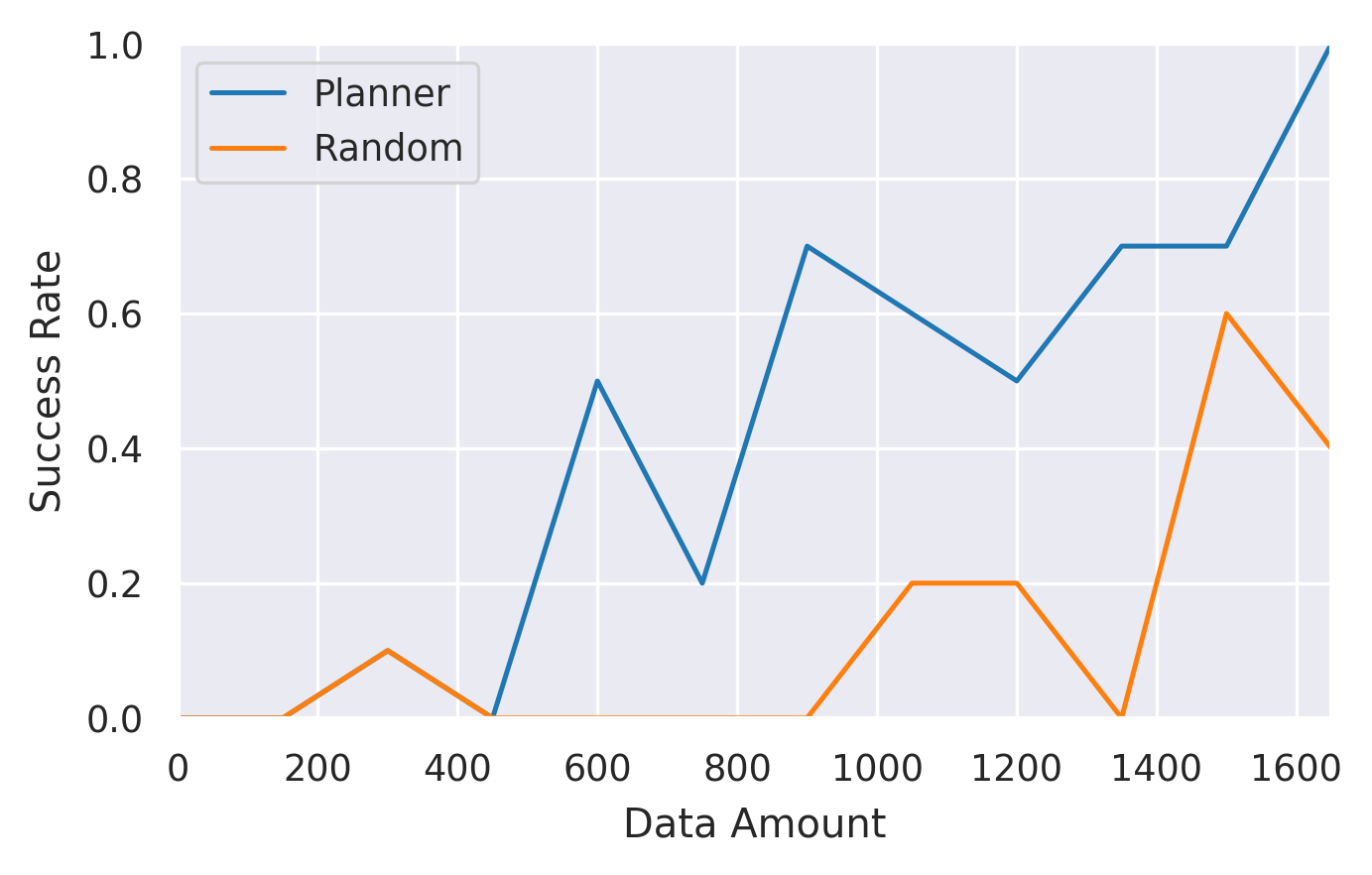}    
    \caption{
        \footnotesize
        % Comparing successes on test task \taskname{B} with only \skillname{Pick and Place} with data generated via random skills vs. using the planner, which is planning for train task \taskname{A}.
        Success on task \taskname{B} with SEM trained on random vs. planner data.
    }
    \vspace{-15pt}
\label{fig:random}
\end{figure}

\textbf{Planning with a Simulator.}
To highlight the need for learning SEMs instead of simulating skill effects for task planning, we compare their planning times in Table~\ref{tab:learn_model_vs_sim_quantitative}.
We only benchmarked cases where the tasks are about moving all blocks and all skills are available.
On average, using the learned model takes less than a second while using the simulator takes ten minutes to half an hour.
Note that these results leverage the simulator's ability to simulate many skill executions concurrently.
Thus, using the simulator for more complex scenarios is prohibitively time consuming due to 1) the large branching factor and 2) a skill's extended horizon, which is much longer than single-step low-level actions or short-horizon motion primitives.
Additionally, Table~\ref{tab:sem_plan_time} shows the plan times for SEMs with increasing number of skills. 
In all cases our planner find plans in less than half a minute. 
% We also see that skills affect multiple tasks differently, \emph{e.g.}, adding the Tray-Sweep skill for Task A makes the task very simple since now it only requires 1 sweep motion. 
% However, the same strategy does not work for Task B (Red blocks in bin) and hence the plan time increases slightly.

\textbf{Training on Planning Data vs. Random Data.}
To evaluate the benefits of using planning data for the iterative training of SEMs, we compare the test-task success rate between our approach and one that generates data by executing random skill sequences.
See results in Figure~\ref{fig:random}.
Training on planning data achieves higher success rates using fewer samples than training on random data does, illustrating the benefit of guiding data collection using a planner.

\begin{table}[!t]
    \centering
    \vspace{5pt}
    \begin{tabular}{l|ll}
                  & \footnotesize{Success} & \footnotesize{Cost} \\ \hline
    \footnotesize{Pick and Place} & 1.0     & 6.68 (0.3) \\
    \footnotesize{+Tray Slide}   & 0.9     & 3.9 (0.9) \\
    \footnotesize{+Tray Sweep}   & 0.8     & 2.61 (0.7)
    \end{tabular}
    \caption{\footnotesize
        Real-world results on \taskname{Red Blocks to Bin}.
        Costs: mean (std).
    }
    \label{tab:real}
    \vspace{-20pt}
\end{table}

\textbf{Real-world Results.}
We built our task domain in the real world (test tasks in Figure~\ref{fig:method}) and used the learned SEMs to plan for the test task \taskname{B}.
Three sets of planning experiments were performed, one with only \texttt{Pick and Place}, one with the addition of \skillname{Tray Slide}, and one with the addition of \skillname{Tray Sweep}.
% We did not implement \skillname{Bin Tilt} in the real world.
Each set of experiments in Table~\ref{tab:real} consists of $10$ planning trials with different initial block configurations.
These results are similar to the ones shown in the task \taskname{A} test curves in Figure~\ref{fig:costs_all_tasks}.
The differences are due to the small changes in real-world object locations and controller implementations.
While we did not fine-tune SEMs on real-world data, doing so may improve real-world performance.

% Franka control using~\cite{zhang2020modular}
\section{Conclusion}
\label{sec:conclusion}
We propose using search-based task planning with learned skill effect models (SEMs) for lifelong robotic manipulation.
Our approach relaxes prior works' assumptions on skill and task representations, enabling planning with more diverse skills and solving new tasks over time.
Using SEMs improves planning speed, while the proposed iterative training scheme efficiently collects relevant data for training.

In future work, we will scale our method to larger number of skills and parameters by using partial expansions and learned parameter samplers. 
We will also explore estimating model uncertainty, using that to both steer planning away from uncertain regions and also fine-tune existing SEMs only on data about which the models are sufficiently uncertain.
\section{ACKNOWLEDGMENT}
\scriptsize{
The authors thank Kevin Zhang for assistance on real-world experiments.
This work is supported by NSF Grants No. DGE 1745016, IIS-1956163, and CMMI-1925130, the ONR Grant No. N00014-18-1-2775, ARL grant W911NF-18-2-0218 as part of the A2I2 program, and Nvidia NVAIL.
}


\clearpage
\balance
\bibliographystyle{IEEEtran}
\bibliography{references}

\clearpage
\begin{appendices}
\normalsize

\section{Related Works}
\label{app:rw}

\begin{table*}[!h]
\centering
\begin{tabular}{l|lllll|lll}
& \multicolumn{5}{l|}{Low-level Skill}                  & \multicolumn{3}{l}{High-level Planner} \\
Paper                         & Policy  & PC       & Effects  & Parameters & PS & Type      & Init Plan & Heuristics    \\ \hline
\cite{kim2017parts}           & MP      & n/a      & Sim      & n/a        & S      & MHA*      & n/a        & H             \\
\cite{song2019multi}          & MP      & n/a      & Sim      & n/a        & LE     & MCTS      & L          & H             \\
% \cite{kim2019pomhdp}          & MP      & n/a      & Sim    & n/a        & S      & RTDP-Bel  & n/a        & H             \\
\cite{pan2020decision}        & MB      & n/a      & Sim      & H, N-SG    & S      & Greedy    & n/a        & H           \\
\cite{lu2020reset}            & MF      & n/a      & L        & n/a        & S      & MPPI      & n/a        & H             \\
\cite{nasiriany2019planning}  & MF      & L        & SG       & L, SG      & LE     & CEM       & H          & H           \\
\cite{xie2020skill}           & MF      & n/a      & L        & L, N-SG    & LE     & CEM       & H          & L             \\
\cite{bagaria2020skill}       & MF      & L        & SG       & n/a        & S      & RRT       & n/a        & n/a           \\
\cite{mandlekar2020iris}      & IL      & n/a      & SG       & L, SG      & S      & RL        & n/a        & L             \\
\cite{ichter2020broadly}      & IL      & n/a      & SG       & L, SG      & S      & Sampling  & L          & n/a           \\
\cite{li2021planning}         & IL      & n/a      & L        & L, N-SG    & LE     & MPC       & n/a        & H             \\
\cite{butzke2014state}        & HC      & H        & H        & n/a        & S      & WA*       & n/a        & H             \\
\cite{simeonov2020long}       & HC      & H        & SG       & L, SG      & V      & Sampling  & H          & n/a           \\
\cite{xu2020deep}             & HC      & L        & L        & H, N-SG    & LE     & MPC       & L          & H             \\
\cite{kaelbling2013integrated}& HC      & H        & H        & H, N-SG    & S, Sym & HPN       & n/a        & H             \\
\cite{konidaris2018skills}    & HC      & L        & L        & n/a        & Sym    & mGPT      & n/a        & n/a          \\
\cite{wang2021learning}       & HC      & L        & SG       & H, N-SG    & S, Sym & PDDLStream      & n/a       & n/a          \\
Ours                          & HC      & H        & L        & H, SG, N-SG  & S      & WA*       & n/a        & H
\end{tabular}

\caption{
Related works on task planning with skills.
}
\label{tab:app-rw}
% \vspace{-32pt}
\end{table*}

Table~\ref{tab:app-rw} compares the discussed related works and our proposed approach.
We include this table to illustrate the subtle but important similarities and differences among prior approaches, which may be overlooked when they are compared at a more general level.
This is not an exhaustive list of all works in relevant areas.
See below for the list of acronyms of each column.
While we list the skill policy and preconditions as hardcoded in our work, this is not a requirement, as our task planner does not restrict the specific implementations of skills.

\textbf{Policy}: 
\textbf{MB} - Model-Based Optimization,
\textbf{MF} - Model-Free RL,
\textbf{MP} - Motion Primitives,
\textbf{HC} - Hardcoded Controllers,
\textbf{IL} - Learned via Imitation Learning

\textbf{Preconditions (PC)}:
\textbf{H} - Hardcoded,
\textbf{L} - Learned

\textbf{Effects}: 
\textbf{H} - Hardcoded,
\textbf{Sim} - Simulator,
\textbf{L} - Learned (usually from simulator data),
\textbf{SG} - Assumes skill reaches subgoal,
\textbf{FC} - Fixed by algorithmic skill construction

\textbf{Parameters}: 
\textbf{H} - Hardcoded ,
\textbf{L} - Learned (e.g. training a CVAE parameter sampler),
\textbf{SG} - Subgoal,
\textbf{N-SG} - Not Subgoals

\textbf{Planning State (PS)}: 
\textbf{S} - State (e.g. low-dimensional object features),
\textbf{V} - Visual (e.g. images, point clouds),
\textbf{LE} - Latent Embeddings (usually learned to encode high-dimensional visual observations),
\textbf{Sym} - Symbols

\textbf{Init Plan} (if the planner uses an initial plan, where does it come from?): 
\textbf{L} - Learned (e.g. from experience or predefined plan skeleton),
\textbf{H} - Hardcoded (e.g. a plan skeleton of a skill sequence, linear interpolation of subgoals, or randomly sampled subgoals)

\textbf{Heuristics} (e.g. task value functions or shaped rewards/costs that guide planning towards a goal): 
\textbf{H} - Hardcoded,
\textbf{L} - Learned (could be from experience, analytical models, or demonstrations)

% Another popular approach to model skills is based on Dynamic Movement Primitives \cite{ijspeert2013dynamical, schaal2006dynamic}, which uses a second order ODE with learned attractors to model both goal-specific as well as general movement patterns. 
% Although prior works have looked at the problem of learning DMP parameters from demonstrations \cite{kober2012reinforcement, kober2014policy}, most works that compose DMPs mostly focus on learning based techniques \cite{lioutikov2016learning, manschitz2015learning}. 
% Since our work does not require any skill-specific assumption it is also possible to use DMP skills with our proposed framework.

\section{Additional Experimental Details}
\label{app:exp-details}

\subsection{SEM Implementation}
\label{app:sem_1}
The GNN model contains four learnable multilayer perceptron (MLP) modules.
The first is node feature embedding module, with two layers of sizes $[32, 32]$.
The second is a message embedding module that pass messages among nodes. 
It has three layers of sizes $[128, 128, 128]$.
The third is a node-level prediction module, with two layers of sizes $[64, S + 32]$, where $S$ is the dimension of per-object features, and the extra $32$ is used to produce graph-level predictions.
The fourth is a graph-level prediction module, which takes as input the sum of the last $32$ dimensions of node-level predictions and passes it through an MLP with sizes $[32, 1]$ to predict skill execution costs.

The loss weight for terminal state prediction is $\lambda_s = 100$, and the weight for cost prediction is $\lambda_c = 1$.
All non-linearities are ReLU.
The network is trained with the Adam optimizer with a batchsize of $128$, initial learning rate of $0.01$, and for $300$ epochs on every training run.

Each input object feature contains the position of the block, its color, and its index.
Index is needed for SEMs to identify which block is being grasped for \skillname{Pick-Place}.
Color and indices are encoded with an offset positional encoding.
For example, for the $i$th color or index, its encoding is a two-dimensional feature vector $[i, i+1]$.
We found in practice that this simple approach allowed our network to capture indices well and also allowed scaling to different number of objects and colors, which cannot be done with one-hot encoding. 

\begin{figure*}[!t]
    \centering
    \includegraphics[width=0.9\linewidth]{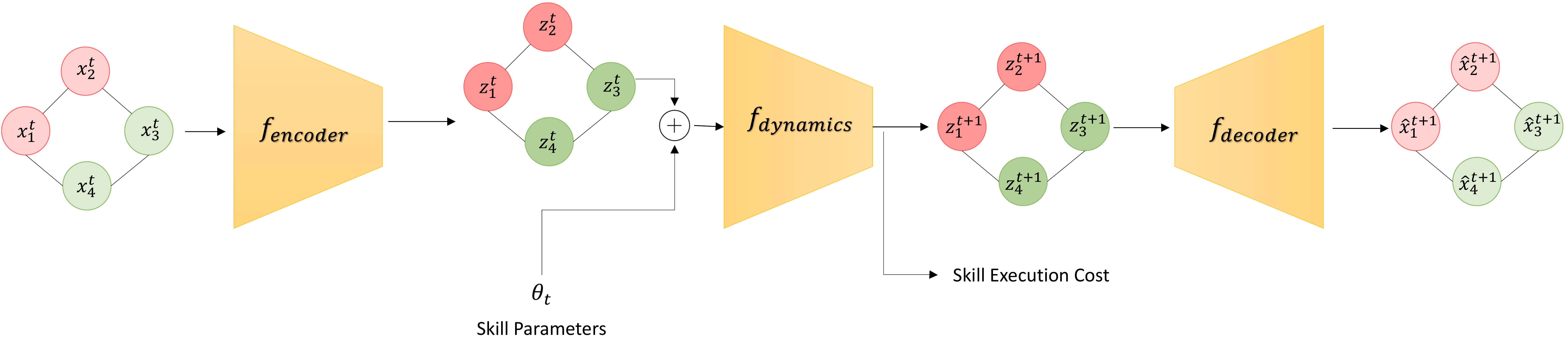}
    \caption{
    Latent Space Model Architecture. The encoder model is used to encode states into latent states while the decoder model converts latent states into original states. While the skill dynamics model acts on the latent state and given skill parameters outputs the final latent state and a skill execution cost. We use a shared encoder-decoder model for all skills while each skill has a separate dynamics model.
    }
    \label{fig:latent_gnn_model}
\end{figure*}

\subsection{Iterative Training}

Instead of collecting all the data from the planner for training the SEM models, we use heuristics to sample the more relevant data.
First, we bias data collection towards longer paths in the planning tree since these paths are more likely to be closer to the relevant task.
Second, we also bias data collection towards newly added skills. 
Since in the beginning we would not have sufficient data for the newly added skill, 
this ensures that we get sufficient data for such new new skills.
Algorithm~\ref{alg:iterative_training} lists the pseudo-code for the data-collection procedure including the heuristics used to sample paths from the planner.

\begin{algorithm}[H]
\caption{
    SEM iterative training pseudocode.
}
\label{alg:iterative_training}

\begin{algorithmic}
    \renewcommand{\algorithmicrequire}{\textbf{Input:}}
    \renewcommand{\algorithmicensure}{\textbf{Output:}}
    \REQUIRE Set of skills $o$s, set of training tasks $\tau$s.
    \FORALL{new skills $o$}
        \STATE Sample $M_0$ initial states from $x_0 \sim p(x_0)$.
        \STATE Sample $P_0$ parameters $\theta$ for each initial state that satisfies $\beta_o(x_0, \theta)$.
        \STATE Simulate $\pi_o$ with all $M_0\times P_0$ (state, parameter) tuples.
        \STATE Add to dataset: $\mathcal{D}_o \leftarrow \{(x_0, \theta, x_T, c_o)\}$.
    \ENDFOR
    \FORALL{skills $o$}
        \STATE Train on $D_o$ for $E$ epochs the skill's SEM from scratch or fine-tune previous model if it exists.
    \ENDFOR
    \FORALL{training tasks $\tau$}
        \STATE Sample $M_p$ initial states from $x_0\sim p(x_0)$.
        \FORALL{sampled initial state $x_0$}
            \STATE $G\leftarrow$ get planning graph by running WA* on $\tau$ with max search depth $N_d$, max node expansions $N_e$, and timeout $T_p$.
            \STATE Sample $N_l$ nodes in $G$'s open list.
            \STATE For each of the $N_l$ nodes, trace their optimal path found so far $P$ from $x_0$.
            \STATE Give each of the $N_l$ paths a weight $w = n_o + 10n_s$, where $n_o$ is the number of old skills in the path, $n_s$ the number of new skills.
            \STATE Sample without replacement $N_s$ paths from all $N_l$ paths with their normalized weights as likelihoods.
            \STATE Simulate each $N_s$ path.
            \STATE Add skill transitions to the corresponding dataset $D_o$ while ignoring duplicates (e.g. some paths may share identical initial segments).
        \ENDFOR
    \ENDFOR
\end{algorithmic}

\end{algorithm}

\section{Weighted A* Planner}
\label{app:planner}
Weight A* (WA*) is a best-first graph search algorithm that expands nodes in the order of lowest $g(x) + \epsilon h(x)$, where $g(x)$ is the total cost of the current optimal path from the initial node to $x$, and $h(x)$ is a heuristic function.
The hyperparameter $\epsilon$ is called the inflation factor and is usually greater than or equal to $1$.
If it is set to $1$, then the search is no different from A*.
The higher the $\epsilon$ above $1$, the greedier the search becomes at following the heuristic.

\subsection{Hyperparameters.}
For all blocks in bin tasks (tasks \taskname{A} and \taskname{C}) with just the \skillname{Pick-Place} skill we use a high value of $\epsilon=20$ since there are not many ways to achieve the task. 
With the addition of more skills i.e, \skillname{Tray-Slide}, \skillname{Tray-Sweep} and \skillname{Bin-Tilt} we set $\epsilon=2$. 
We found that this value to be sufficient to choose optimal plans. 
Additionally we also set the maximum search depth for both tasks \taskname{A} and \taskname{C} to be 8, since the longest plan for these tasks should be of length 8.
For the colored blocks in bin tasks (tasks \taskname{B} and \taskname{D}), we used $\epsilon=2$ and a max search depth of $5$.
Additionally, we sample a larger number of parameters for \skillname{Pick-Place} skill as compared to the other skills. 
This is because the \skillname{Pick-Place} skill affects each block separately and thus to find optimal plans we need to sample sufficient parameters for each block. 
This necessitates a larger number of parameters for this skill. 
For \skillname{Pick-Place} we sample 24 parameters while for all other skills we sample $[4, 6]$ parameters.

\subsection{Guarantees on the Constructed Graph}
WA* guarantees completeness and bounded suboptimality on a given graph.
Here we show that under smoothness assumptions, the graph constructed with our parameter sampling approach is suitable for search-based planning.
Theorem~\ref{th:goal} is about completeness --- the distance between the reached goal state by any solution path and the closest last state of a path on the graph is bounded.
Theorem~\ref{th:cost} is about solution quality --- for any solution path with a certain cost, the graph will contain a solution path with a bounded cost difference.

\begin{definition}
The \textit{dispersion}~\cite{lavalle2006planning}  of a finite set
A of samples in a metric space  $(X, \rho)$ is defined as
$$\delta(A) = \sup_{x \in X} \min_{p \in A} \rho(x, p)$$
Intuitively, it is  the radius of the largest empty ball that can be drawn around any point in $X$ without intersecting any point in A.
\end{definition}

In the following discussion, we assume all preconditions and the goal set are open sets.
\begin{theorem}
\label{th:goal}
Let the skill transition function $f_o: \mathcal{X} \times \Theta \rightarrow \mathcal{X}$ be Lipschitz continuous  with a maximum Lipschitz constant $K$ and $\eta = (x_0,o_0,\theta_0,x_1,\cdots,x_N)$ be a solution to the planning problem.
Then, $\exists$ a  path $\eta' = (x_0,o_0,\theta_0',x_1'\cdots,x_N')$ on the constructed search graph such that $||x_N - x_N'|| \leq 2 \delta \kappa_N = 2 \delta \frac{K(K^N - 1)}{K - 1}$ if the dispersion $\delta$ of parameter samples is small enough.
\end{theorem}

\textbf{Proof} Consider an instance of the randomly generated search tree $\mathcal{T}$ of depth more than $N$.
We pick a small enough $\delta$ such that $\mathcal{T}$ contains at least one path $\eta'$ that has the same sequence of skills as $\eta$.
However, due to random sampling, the sampled parameters may not be the same.
For every $\theta_i$ at a state, $\exists$ a parameter  sample $\theta_i'$  such that
\begin{equation}
    ||\theta_i - \theta_i'|| \leq 2\delta
    \label{eq:dispersion}
\end{equation}
Using the definition of Lipschitz continuity and the triangle inequality, we have
\begin{align}
||f_{o_i}(x_{i}, \theta_i) - f_{o_i}(x_{i}', \theta_i')|| 
&\leq K||(x_{i}, \theta_i) - (x_{i}', \theta_i')|| \\
&\leq K||x_{i} - x_{i}'|| + K||\theta_i - \theta_i'||
\end{align}
In particular,
$$||x_1 - x_1'|| = ||f_{o_0}(x_0, \theta_0) - f_{o_0}(x_0, \theta_0')|| \leq K||\theta_0 - \theta_0'||$$
and
\begin{equation}
\begin{split}
||x_N - x_N'|| &= ||f_{o_{N-1}}(x_{N-1}, \theta_{N-1}) - f_{o_{N-1}}(x_{N-1}', \theta_{N-1}'|| \\
               &\leq K||\theta_{N-1} - \theta_{N-1}'|| + K||x_{N-1} - x_{N - 1}'|| \\
               &\leq \sum_{i=0}^{N-1}{K^{N-i}||\theta_i - \theta_i'||} \leq \Delta \theta \sum_{i=0}^{N-1}{K^{N-i}}\\
               &\leq 2\kappa_N \delta \text{  (Using inequality \ref{eq:dispersion})}
               \label{eq:state_bound}
\end{split}
\end{equation}
where $\kappa_N = \frac{K(K^N - 1)}{K - 1}$.

To ensure that we have a path $\eta'$ on the graph that terminates $\epsilon$-close to the terminal state of $\eta$, we require that $||x_N - x_N'|| \leq 2 \kappa_N \delta \leq \epsilon $.

To guarantee that $\eta'$ will also be a solution, we additionally need to ensure that $\epsilon < r$, where $r$ is the radius of the largest ball we can draw around $x_N$ in $G$ (goal set). 
This is always possible if $G$ is an open set.

\begin{theorem}
Let the cost function be Lipschitz continuous with a maximum Lipschitz constant of $L$.
Let $\eta$ be a solution path of cost $c(\eta)$ and $r > 0$, then for a sufficiently small dispersion $\delta$, $\exists$ a solution path $\eta'$ on the constructed search graph with cost $c(\eta') \leq c(\eta) + \delta NL[1 + \sum_{i=1}^N \kappa_i]$.
\label{th:cost}
\end{theorem}

\textbf{Proof} 
Choose $\delta$ such that $2\kappa_N\delta < r$ and $\exists$ a path $\eta'$ on the graph with the same sequence of skills as $\eta$.
Then, from the previous theorem, $\exists$ a path $\eta'$ on the graph such that $||x_N - x_N'|| \leq 2\kappa_N\delta < r$.
By the definition of $r$, $x_N' \in G$ and hence $\eta'$ is a solution path.
Next, we bound the cost of this path.

The cost of a path $c(\eta) = \sum_{i=0}^{N-1}c_{o_i}(x_i, \theta_i)$.
For convenience, here we write the cost function as skill-specific cost functions that vary w.r.t the initial state and skill parameters, instead of step-wise costs on state and low-level controls.
Let $c_i = c_{o_i}(x_i, \theta_i)$ and  $c_i' = c_{o_i}(x_i', \theta_i')$.
Then, using the definition of Lipschitz continuity, we have
\begin{equation}
\begin{split}
||c_i - c_i'|| &\leq L||(x_i,\theta_i) - (x_i',\theta')||\\
                &\leq L||x_i - x_i'|| + L||\theta_i - \theta_i'|| \\
                &\leq L[2\kappa_i\delta + \delta] 
\end{split}
\end{equation}
Hence,
\begin{equation}
    \begin{split}
        ||c(\eta) - c(\eta')|| &= ||\sum_{i=0}^{N-1}{c_i - c_i'}|| 
                                \leq \sum_{i=0}^{N-1}||c_i - c_i'|| \\
                                &\leq \delta NL[1+ 2\sum_{i=1}^N \kappa_i]
                                \label{eq:cost_bound}
    \end{split}
\end{equation}

The dispersion of a set of sampled parameters depends on the sampling strategy used.
For uniformly random sampling, we can only estimate it probabilistically.
~\cite{hinrichs2020expected} prove bounds on the expected dispersion of a set of i.i.d points which tightens with more points, i.e., we are more confident of a lower expected dispersion with a larger number ($B_o$) of parameter samples.

Theorem~\ref{th:cost} provides two additional observations.
First, skills with dynamics that don't change fast, i.e., their Lipschitz constant are small, can be approximated with a sparse graph (less parameter sampling).
Second, longer horizon tasks require more parameter sampling to guarantee good quality plans.

\section{Task Domain Skill Details}

All skill policies are implemented by end-effector waypoint following, where trajectories between waypoints are computed via min-jerk interpolation, and low-level control is achieved by end-effector impedance control.
A subset of waypoints for each skill is indirectly determined by the skill parameters.
For example, for \skillname{Pick-Place}, the object index parameter determines which object to pick.
Together with the current state, they are used to compute the waypoints associated with the picking motion.

\subsection{Pick-Place}

This skill picks up a chosen block and places it at a target location.

\textbf{Parameter.}
A 4-vector consisting of an index corresponding to which block to pick and a 3D position for placement

\textbf{Parameter Sampling.}
For which block to pick, we sample the index of a block on the table with collision-free grasps.
For the placement position, we first sample which general placement region to use, table, bin, or tray, with probabilities $[0.2, 0.5, 0.4]$.
Then, we randomly sample a position on the surface of reach placement region.
The bin placement region is the half of the bin that is closer to the robot.

\textbf{Precondition.}
The precondition function check returns satisfied if the placement position is collision free.
It assumes the pick grasp is collision free and the placement location is reachable.

\subsection{Tray Slide}

This skill picks up the tray, brings it over the bin, rotates the tray to let blocks fall to the bin, then brings the tray back to its original pose.

\textbf{Parameter.}
One value that determines where along the bin does the slide rotation motion.
This allows dropping blocks over the close or far side of the bin.

\textbf{Parameter Sampling.}
This is done via uniform sampling over a range of the length of the bin.

\textbf{Precondition.}
The precondition is satisfied when there is at least one block on the tray.

\subsection{Tray Sweep}
This skill picks up the tray, rotates it $90^\circ$, brings it down toward the table, sweeps downstream objects into the bin, and returns the tray to its original pose.

\textbf{Parameter.}
One value that determines where along the table the sweep motion begins.

\textbf{Parameter Sampling.}
This is done via uniform sampling over a range of the width of the table.

\textbf{Precondition.}
The precondition is satisfied when there is at least one block that will be swept by the skill and the starting location for the tray is collision-free.

\subsection{Bin Tilt}
This skill lifts one side of the bin by a desired angle, allowing objects to slip down toward the far side of the bin, before return the bin to the original configuration.

\textbf{Parameter.}
One value that determines the angle of the tilt.
With a shallow angle it is possible for some blocks to remain in the near side of the bin, or not move at all due to friction.

\textbf{Parameter Sampling.}
This is done via uniform sampling in the range of $[5^\circ, 20^\circ]$.

\textbf{Precondition.}
The precondition is satisfied if there is at least one block in the bin.

\begin{figure*}[t]
    \centering
    \includegraphics[width=0.9\linewidth]{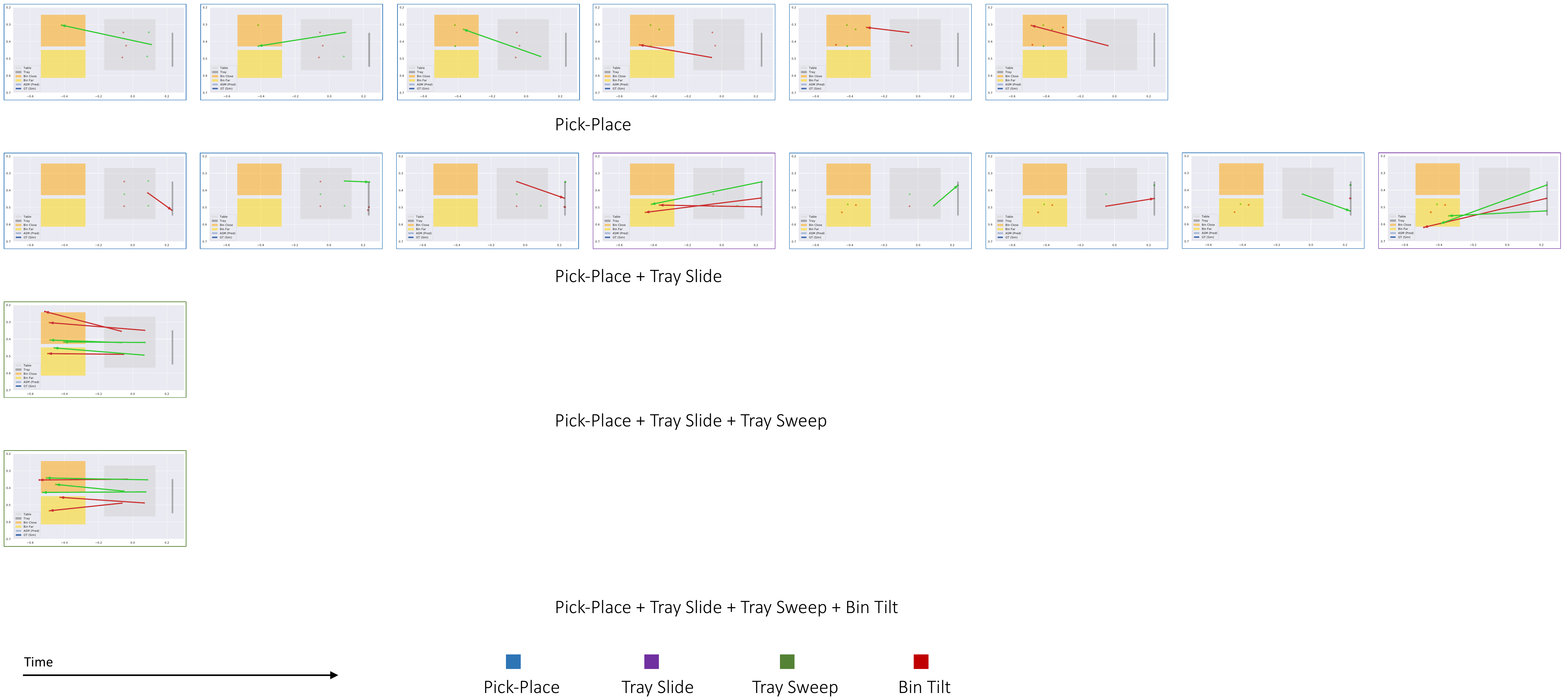}
    \caption{
    Qualitative results: Plans found for \taskname{Task-A} with increasing number of skills. Left most columns are the skills executed at t=0, with skills being executed as we move towards right.
    Each row also lists the different skills used to plan.
    }
    \label{fig:qualitative_task_a}
\end{figure*}

\begin{figure*}[t]
    \centering
    \includegraphics[width=0.9\linewidth]{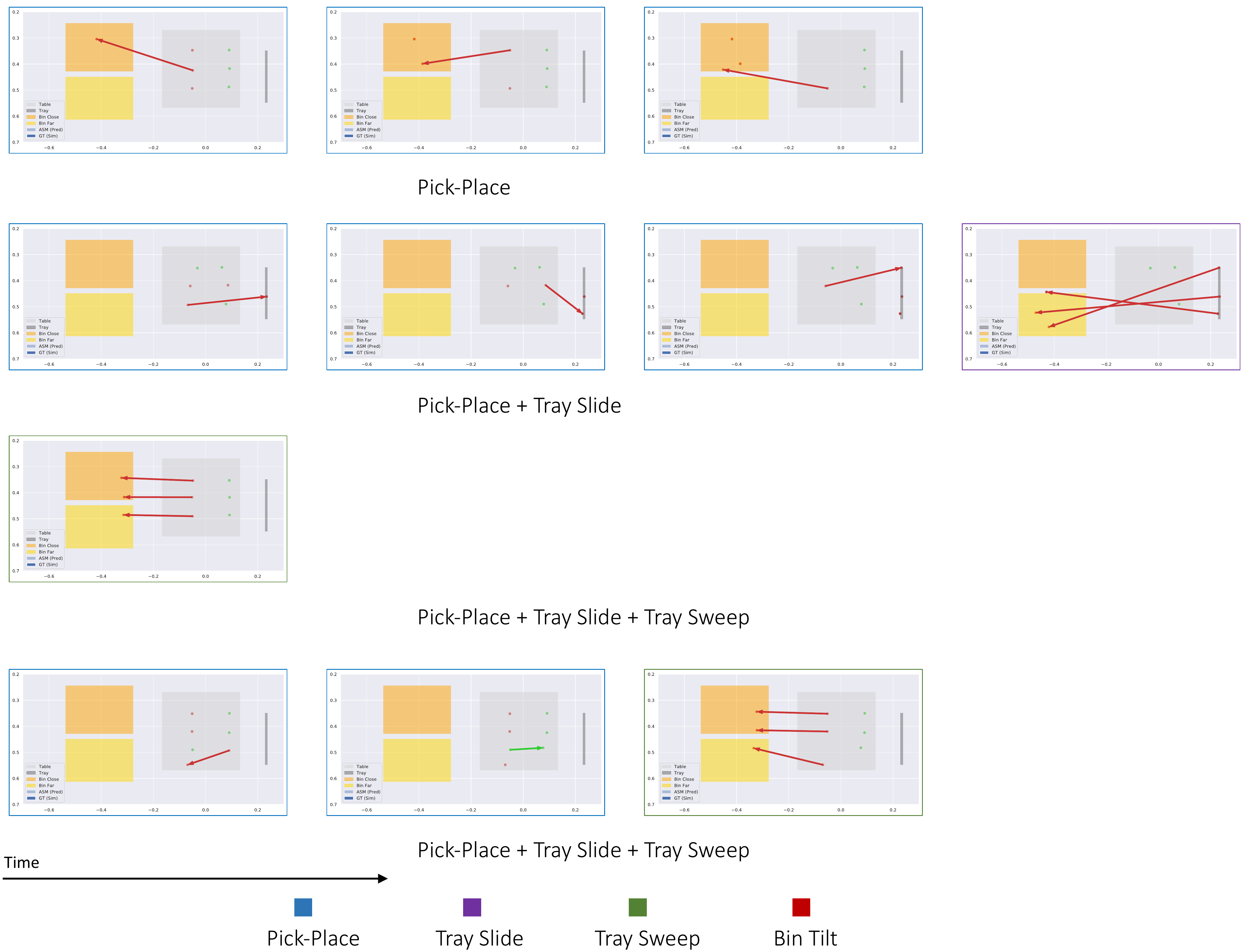}
    \caption{
    Qualitative results: Plans found for Task-B with increasing number of skills. Left most columns are the skills executed at t=0, with skills being executed as we move towards right.
    Each row also lists the different skills used to plan.
    }
    \label{fig:qualitative_task_b}
\end{figure*}

\section{Additional Results}
\label{app:additional-results}

\subsection{Qualitative Planning Results}

In this section we show qualitative results for our train and test tasks. 

Figure~\ref{fig:qualitative_task_a} shows the qualitative results for the train task \taskname{A} with an increasing number of skills. 
In the above result we see that with only \skillname{Pick-Place} skill the planner finds a plan which needs to pick all blocks and place them in the bin.
However, after adding the \skillname{Tray-Slide} skill (row 2) the planner finds a longer plan but with a lower cost \ref{fig:costs_all_tasks}. 
This lower cost is a result of being able to use the Tray-Slide skill to transport multiple blocks to the bin directly.
Additionally, with the addition of \emph{Tray-Sweep} skill we can perform the task with only 1 skill which significantly reduces the overall plan cost.

\begin{figure*}[t]
    \centering
    \includegraphics[width=0.9\linewidth]{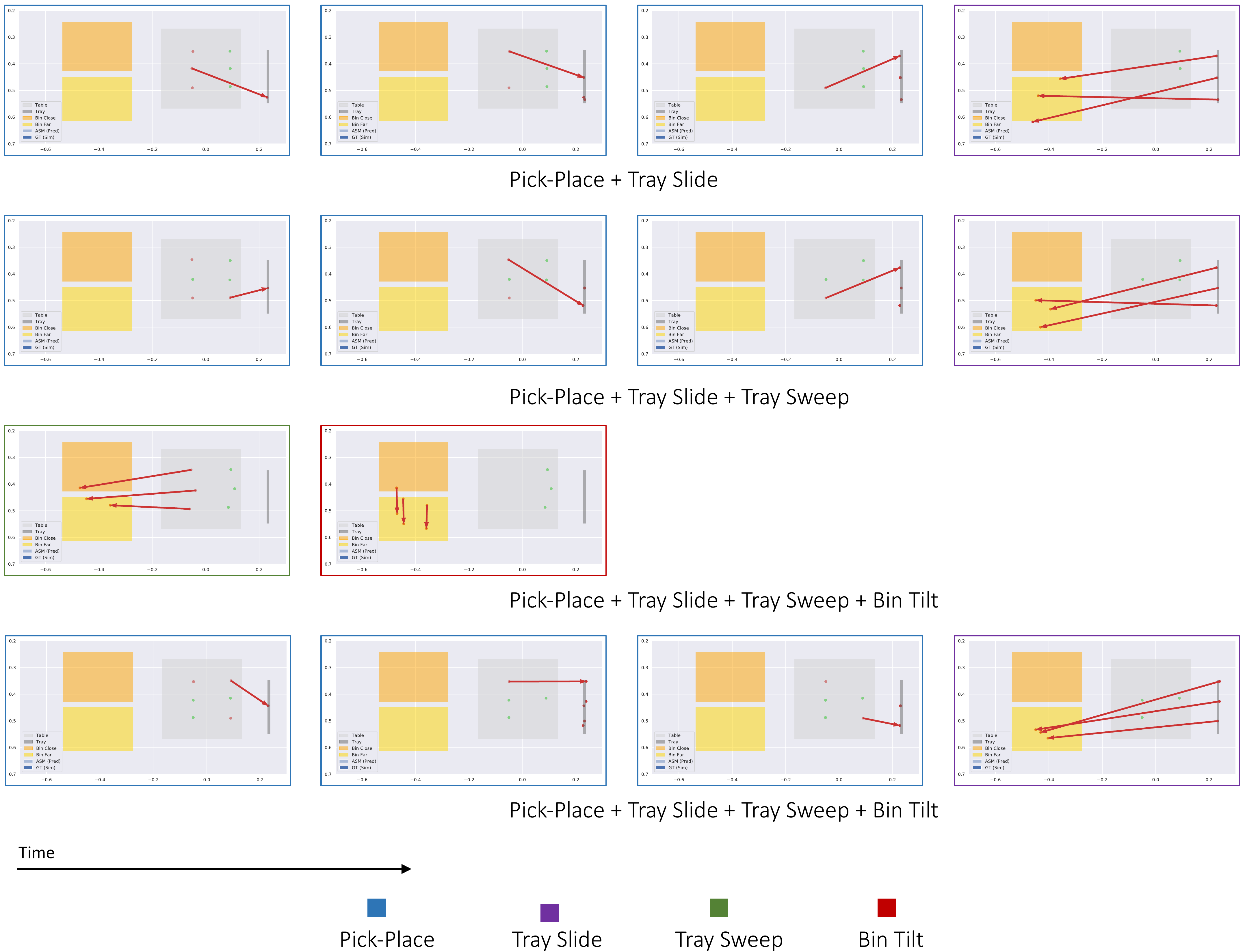}
    \caption{
    Qualitative results: Plans found for \taskname{Task-D} with increasing number of skills. Left most columns are the skills executed at t=0, with skills being executed as we move towards right.
    Each row also lists the different skills used to plan.
    }
    \label{fig:qualitative_task_d}
\end{figure*}

Similarly, Figure~\ref{fig:qualitative_task_b}~and ~\ref{fig:qualitative_task_d} show the qualitative results for both test tasks \taskname{B} and \taskname{D} respectively.
Similar to before, with an increasing number of skills the overall plan and its associated cost changes significantly. 
For task \taskname{B} we observe that adding \skillname{Tray-Sweep} improves the performance only in some scenarios as shown in Figure~\ref{fig:qualitative_task_b} (rows 3 and 4). 
In another instance the robot can also perform \skillname{Pick-Place} to align the red  blocks to sweep them into the bin (row 4). 

Similarly, for test-task \taskname{D} we observe that using \skillname{Pick-Place} alone is not sufficient (as observed in Figure~\ref{fig:costs_all_tasks}).
However, adding the \skillname{Tray-Slide} skill makes the task feasible (top row). 
However, in contrast to tasks \taskname{A} and \taskname{B}, adding \skillname{Tray-Sweep} does not affect task execution since the robot cannot move the blocks to the far bin with this skill. 
We observe this in Figure~\ref{fig:qualitative_task_d} where the top two rows use the same set of skills.
However, adding the \skillname{Bin-Tilt} skill allows the planner to use the \skillname{Tray-Sweep} skill, 
as shown in Figure~\ref{fig:qualitative_task_d} (row 3, second row from bottom).

\subsection{Video Results Link}
Video results for most tasks and settings can be found at the project page \url{https://sites.google.com/view/sem-for-lifelong-manipulation}.

\subsection{Failure Modes}

Here we discuss examples of planning and execution failures due to insufficiently trained SEMs.

First, if SEM predictions are not sufficiently accurate, then the predicted states would do not satisfy the preconditions of necessary skills, and the planner will not find a plan. 
For instance, when \skillname{Tray-Sweep} is trained with less data, the learned SEM predictions can incorrectly predict that the blocks move to the edge of the table instead of being moved to the bin.
Thus, the planner considers this to be a non-optimal action and instead resorts to using the remaining skills (\skillname{Pick-Place} and \skillname{Tray Slide}), which leads to a sub-optimal plan if one is found. 
Another scenario where incorrect SEM predictions lead to failure is when \skillname{Tray-Slide} is added. 
Given the previously used \textbf{Pick-Place} was mostly trained on moving blocks from the table to the bin (because of the train-task bias), the existing \textbf{Pick-Place} SEM may inaccurately predict that the skill is unable to move blocks to the tray.
This results in the failure for \skillname{Tray-Slide} preconditions which finally results in either the planner timing out or finding sub-optimal plans. 

Second, there are also scenarios where the planner is able to find a plan despite inaccurate SEM predictions, but the plan does not work in execution. 
One common instance is with \skillname{Pick-Place}, where early on with insufficient data the SEM often inaccurately predicts that the skill will move blocks to the bin, despite the target placement location being on the edge of the table.
Given the incorrect prediction by the SEM, the planner mistakenly believes that the block has been placed in the bin, and it may find a plan but its execution does not lead to success. 

\subsection{Latent Space SEMs}
\label{app:latent-space-sem}

\begin{figure}
    \centering
    \includegraphics[width=0.9\linewidth]{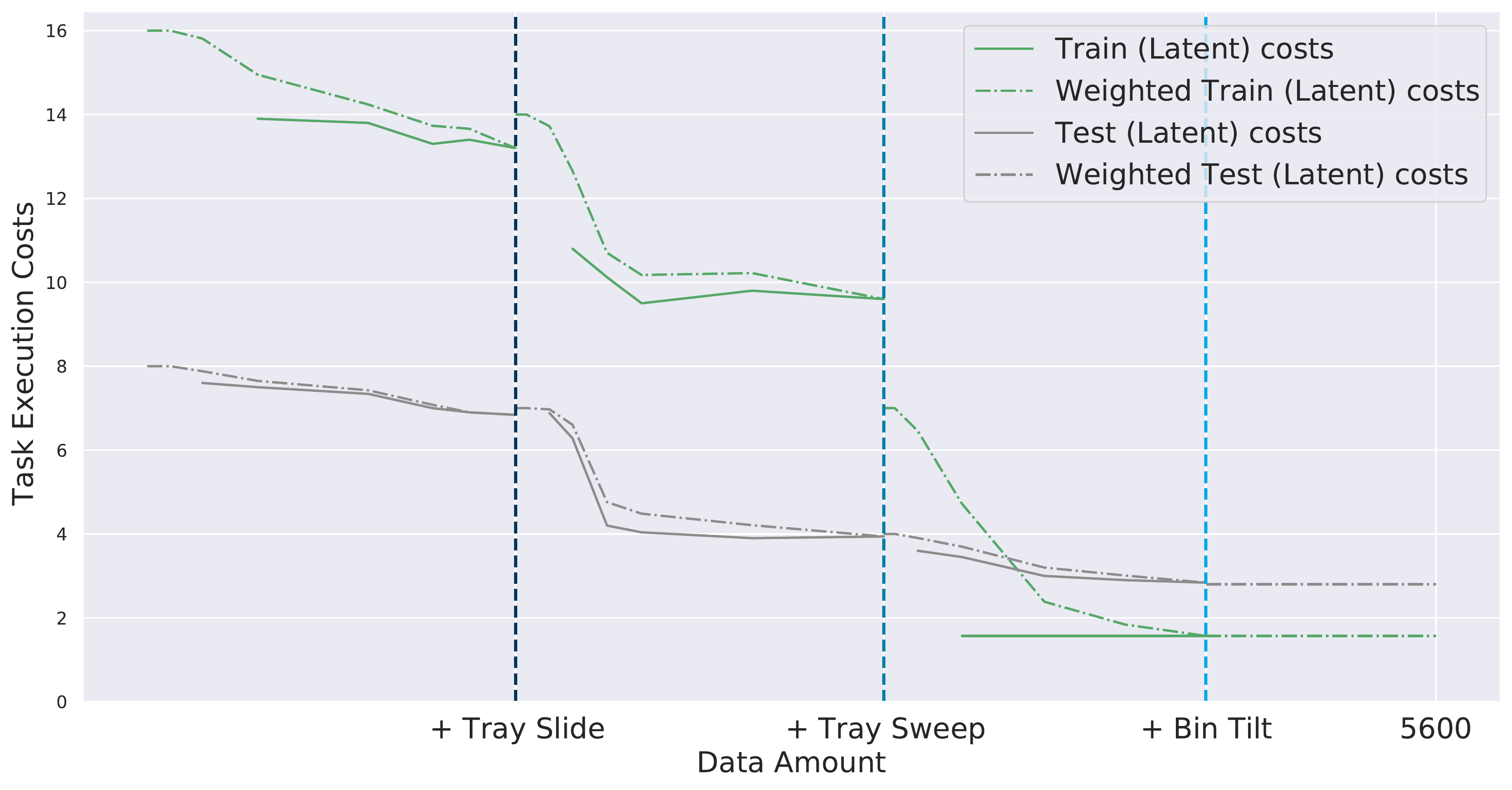}
    \captionof{figure}{
        \footnotesize
        Execution costs for tasks \taskname{A} and \taskname{B} using SEMs in latent-space.
    }
    \label{fig:latent}
\end{figure}

In our main results we trained SEMs directly using the state space for each skill separately. 
We also evaluate how SEMs perform when trained with a shared latent space. 
% A shared encoder and decoder are learned.
% The encoder maps state-space inputs to latent embeddings, and the decoder reconstructs inputs from the latent space.

Our latent space SEMs implementation uses an encoder-decoder architecture as visualized in Figure~\ref{fig:latent_gnn_model}. 
The encoder model $f_{\text{encoder}}$ uses the state as an input and outputs a latent state. 
The dynamics model $f_{\text{dynamics}}$ then uses this latent state along with the skill parameters $\theta_t$ to outputs the next latent state denoted as $\hat{z}^{t+1}_{j}$ in Figure~\ref{fig:latent_gnn_model}. 
Along with the next latent state the skill-specific dynamics model also outputs the skill execution execution cost.
We note that the dynamics model is specific to each skill.
The decoder is used to decode the latent states to the original state space.
Separate latent-space SEMs are learned for each skill. 

Figure~\ref{fig:latent_gnn_model} only visualizes the encoder-decoder acting on state $x^{t}_{j}$ (state at time $t$) and $z^{t+1}_{j}$ (latent state at $t+1$) respectively. 
However, during training the encoder and decoder model are used to reconstruct all states \emph{i.e.} both the current state and the next state.
In our implementation, we use a latent space of size 32.
Additionally all our models \emph{i.e.}, the encoder, decoder and dynamics model for all skills are implemented using the same GNN implementation as discussed in Appendix-\ref{app:sem_1}. 

Figure~\ref{fig:latent} plots the costs when using the latent-space models on the train-test task pair of \taskname{A} and \taskname{B}, which has comparable performance to state-space models (Figure~\ref{fig:costs_all_tasks} left).
This is a promising result since such a latent space could allow us to plan using image inputs; however, we leave image-space planning for future work. 
% Latent-space architecture details are in Appendix-\ref{app:latent-space-sem}.

\begin{table*}[!t]
\centering
\begin{tabular}{l|llll}
Problem & Cost (Ours) & Cost (Random) & Planning Time (Ours) & Planning Time (Random) \\ \hline
1       & 9.0 (0.83)  & 11.63 (0.93)  & 20.2 (7.9)           & 641.90 (375.93)        \\
2       & 1.5 (0.0)   & 2.34 (1.03)   & 0.6 (0.5)            & 5.69 (11.51)           \\
3       & 4.3 (0.21)  & 6.67 (0.62)   & 14.9 (8.2)           & 432.22 (345.76)        \\
4       & 3.89 (0.8)  & 5.72 (1.06)   & 18.0 (14.3)          & 311.78 (337.30)       
\end{tabular}
\caption{
    Planning performance comparison with Random Planner.
    Values are averaged across 10 trials in which the random planner found a plan within 1000s, and stds are in the parenthesis
}
\label{tab:random_planner}
\end{table*}

\subsection{Random Planner}

To demonstrate the need for a guided planner to efficiently search (skill, parameter) tuples, we performed an experiment to compare planning performance with a random planner.
The random planner randomly picks the next node to expand in the search graph, instead of in the order of their f-value (optimal cost to come + weighted heuristic). 
We evaluated the random planner on the following 4 planning problems:

\begin{enumerate}
    \item Task A (all blocks to bin) w/ \skillname{Pick-Place}, \skillname{Tray Slide}
    \item Task A (all blocks to bin) w/ \skillname{Pick-Place}, \skillname{Tray Slide}, and \skillname{Tray Sweep}
    \item Task B (red blocks to bin) w/ \skillname{Pick-Place}, \skillname{Tray Slide}
    \item Task B (red blocks to bin) w/ \skillname{Pick-Place}, \skillname{Tray Slide}, and \skillname{Tray Sweep}
\end{enumerate}

See results in Table~\ref{tab:random_planner}.
In all cases the random planner took an order of magnitude longer to plan on average than a guided planner. 
The achieved costs are also all higher than the costs of the plans found using the guided planner. 
The complexity of the planning problem is caused by 1) we do not assume knowledge of a plan skeleton or the exact number of skills needed to complete the task and 2) the high branching factor (about 30~40 depending on the particular scenario) caused by sampling skill parameters. 

\subsection{Learned Precondition Model}

Here we perform a small experiment to show it is possible for our method to support learned skill precondition models.
These learned skill precondition models take as input a (skill, parameter) tuple and output a binary output which signifies if the given skill can be executed at the current state. 
We use a GNN architecture as our precondition model. 
To collect data for training skill preconditions we initialize the environment to a random configuration by first selecting the block to place and then selecting a place location for it. 
This block can be placed on the table, on the bin or on the tray. 
We then execute the skill with some randomly sampled parameter to note if the skill can be successfully executed in this scenario. 
Using this scheme we collect around 24000 data samples for the Pick-Place skill and 8000 data samples for the Tray-Slide skill. We use this to train our precondition model. 
No task-specific data is used to train these precondition models. 

This learned precondition model is combined with the learned SEMs to plan on test task \taskname{B} (red blocks to bin). 
We execute these learned models for 20 different scenarios with 3 seeds. 
In addition to the average cost, we also report the average success rate since in some scenarios errors in the learned precondition models may affect task success. 

See results in Table~\ref{tab:learned_preconditions}.
The learned precondition models also allow the planner to have a high test-task success ratio, which indicates the effectiveness of the learned precondition models and learned SEMs. 
The precondition models were only trained with random data and their accuracy can be improved by additionally utilizing data incurred from planning for train-tasks. 

\begin{table}[!h]
\centering
\begin{tabular}{l|ll}
Skill                       & Success Rate Mean (std) & Cost \\ \hline
\skillname{Pick-Place}              & 0.96 (0.02)             & 6.44 \\
\skillname{Pick-Place} + \skillname{Tray Slide} & 0.90 (0.04)             & 4.35
\end{tabular}
\caption{
    Planning results for using learned precondition models on Task \taskname{B}.
}
\label{tab:learned_preconditions}
\end{table}

\subsection{Generalization to Object Features}

In the experiment domain the state space includes the 3D position of each block, its color, and its index. 
Here we run two small experiments to investigate how well the learned SEMs would generalize to variations in object features not in the state space that have varying impacts on skill effects.

\textbf{Generalization to object sizes}: In the first experiment, we show that trained SEMs can generalize to objects of different sizes. 
We show this using \skillname{Pick-Place} and \skillname{Tray Slide}. 
Both skills have been trained on square objects of size 4 cm and we show that these models do generalize to objects of size 2cm and 3cm. We see similar quantitative performance on these different sized objects. 
See Table~\ref{tab:obj_size} for the average costs for different sized blocks. 
The trained SEMs do generalize well to objects of different sizes. 

\begin{table}[!h]
\centering
\begin{tabular}{l|lll}
Skill                       & 4cm  & 3cm  & 2cm  \\ \hline
Pick-Place              & 6.41 & 6.56 & 6.47 \\
Pick-Place + Tray Slide & 4.1  & 4.04 & 4.12
\end{tabular}
\caption{
    Average cost for Test-Task \taskname{B} showing generalization of SEMs to different object sizes
}
\label{tab:obj_size}
\end{table}

\textbf{Generalization to object counts}: In the second experiment we show that since object color is a property included in the state representation our learned SEMs can also generalize to different numbers of colored objects. 
During training we always use 3 red colored blocks and 3 green colored blocks. 
We evaluate these learned SEMs for Test-task B and on scenes with 1 red colored block and 5 green colored blocks as well as 2 red colored blocks and 4 green colored blocks.
We initially use Pick-Place skill only and then incrementally add the Tray-Slide skill. 
See Table~\ref{tab:obj_count}, which shows that the learned SEMs are able to generalize well to these different settings. 

\begin{table}[!h]
\centering
\begin{tabular}{lll}
\multicolumn{1}{l|}{Skill}                       & Scene        & Cost \\ \hline
\multicolumn{1}{l|}{Pick-Place}              & 1 Red Block  & 1.91 \\
\multicolumn{1}{l|}{Pick-Place}              & 2 Red Blocks & 4.33 \\
\multicolumn{1}{l|}{Pick-Place + Tray Slide} & 1 Red Block  & 1.96 \\
\multicolumn{1}{l|}{Pick-Place + Tray Slide} & 2 Red Blocks & 3.20
\end{tabular}
\caption{
    Average cost for Test-Task \taskname{B} showing generalization of SEMs to different numbers of objects.
}
\label{tab:obj_count}
\end{table}

\end{appendices}

\end{document}